\documentclass[10pt,twocolumn,letterpaper]{article}

\usepackage[pagenumbers]{cvpr}      
\usepackage{placeins}
\usepackage{tcolorbox}
\definecolor{cvprblue}{rgb}{0.21,0.49,0.74}
\usepackage{multirow} 
\usepackage[pagebackref,breaklinks,colorlinks,allcolors=cvprblue]{hyperref}

\def\paperID{} 
\def\confName{CVPR}
\def\confYear{2026}

\title{Understanding Counting Mechanisms in\\
Large Language and Vision-Language Models}

\author{
  \text{Hosein Hasani}\thanks{Equal contribution.},
  \text{Amirmohammad Izadi}\textsuperscript{*},
  \text{Fatemeh Askari}\textsuperscript{*},
  \text{Mobin Bagherian}\textsuperscript{*}, \\
  \vspace{3mm} 
  \text{Sadegh Mohammadian},
  \text{Mohammad Izadi},
  \text{and Mahdieh Soleymani Baghshah}\\
  \vspace{1mm} 
  Sharif University of Technology \\
{\tt \small
\{hosein.hasani, amirmohammad.izadi01, fatemeh.askari79, mobin.bagherian01\}@sharif.edu}\\
{\tt \small
\{sadegh.mohammadian01, izadi, soleymani\}@sharif.edu
}
}

\begin{document}
\maketitle
\begin{abstract}

Counting is one of the fundamental abilities of large language models (LLMs) and large vision-language models (LVLMs). This paper examines how these foundation models represent and compute numerical information in counting tasks. We use controlled experiments with repeated textual and visual items and analyze counting in LLMs and LVLMs through a set of behavioral, observational, and causal mediation analyses. To this end, we design a specialized tool, CountScope, for the mechanistic interpretability of numerical content. Results show that individual tokens or visual features encode latent positional count information that can be extracted and transferred across contexts. Layerwise analyses reveal a progressive emergence of numerical representations, with lower layers encoding small counts and higher layers representing larger ones. We identify an internal counter mechanism that updates with each item, stored mainly in the final token or region. In LVLMs, numerical information also appears in visual embeddings, shifting between background and foreground regions depending on spatial composition.
We further reveal that models rely on structural cues such as separators in text, which act as shortcuts for tracking item counts and strongly influence the accuracy of numerical predictions. Overall, counting emerges as a structured, layerwise process in LLMs and follows the same general pattern in LVLMs, shaped by the properties of the vision encoder.
\footnote{Code available at: \url{https://github.com/sharif-ml-lab/counting-mechanisms}}
\end{abstract}    
\section{Introduction}
\label{sec:introduction}

   


Counting is a fundamental operation that supports reasoning, perception, and symbolic understanding~\cite{feigenson2004core,dehaene2011numbersense}. For language and vision-language models, the ability to count underlies a wide range of tasks, from object enumeration and quantitative description to basic arithmetic reasoning \cite{paiss2023clipcount, campbell2024binding}. Research has shown that LLMs often struggle with simple counting tasks, exhibiting failures related to repetition tracking, token boundaries, and prompt variations
\cite{fu2024letters, zhang2024tokenization, canwecount2024}. Prior work has mainly assessed counting through behavioral accuracy or prompt-based evaluations \cite{paiss2023clipcount, rahmanzadehgervi2024blind, campbell2024binding, canwecount2024, zhang2024tokenization}. Despite this progress, there is still limited work that directly investigates the internal mechanisms underlying counting in large language models (LLMs) and large vision-language models (LVLMs).

At the same time, recent advances in mechanistic interpretability have provided tools to study where and how task-dependent information is represented \cite{vig2020cma, stolfo2023cma, ghandeharioun2024patchscopes}. In this work, we seek to uncover the mechanisms by which these models represent, update, and output numerical information during counting tasks. Rather than focusing on behavioral accuracy alone, our goal is to identify the specific internal components that mediate counting and to establish how numerical representations emerge across layers and modalities. By combining causal and observational analyses, we provide a systematic characterization of counting, tracing how latent numerical signals are encoded, transferred, and decoded within the model.

Our approach builds on mechanistic interpretability techniques that allow direct intervention in model activations to test their causal role in generating numerical outputs.
We design a controlled experimental framework that presents repeated objects (either textual or visual items) and examines how models infer their number.
Using a combination of causal mediation analysis and online and offline activation patching, we isolate where and how count-related information is stored within the model. In particular, we design a specialized tool, \textit{CountScope}, that causally reveals count-related information in internal activations.

Our analyses on multiple LLMs and LVLMs show that numerical signals arise early in the context, concentrate in the final items, and follow a consistent latent trajectory across layers. We find that models maintain position-dependent representations that often rely on separator structure as a shortcut. We causally identify an internal counter mechanism that performs numerical computation across tokens and layers. Based on the observation that counting ability is limited by transformer depth, we suggest a practical implication: LVLMs can leverage the capacity of visual representations and distribute counting computation across visual tokens rather than relying solely on transformer layers.
In conclusion, this study aims to present a unified view of how LLMs and LVLMs represent, update, and read out numerical information during counting tasks.
\section{Preliminaries}

\subsection{Definitions}
Mechanistic interpretability aims to uncover how internal computations in LLMs give rise to observed behaviors \cite{wang2023ioi}. 
Activation patching provides a practical framework for this purpose, offering causal rather than purely correlational analysis~\cite{vig2020cma, zhang2024activationpatching, heimersheim2024activationpatching}. 
By directly intervening in intermediate activations and observing resulting output changes, these methods allow researchers to identify which internal components causally mediate specific functions.
Common patching types include zero patching, which removes contributions by setting activations to zero; mean patching, which replaces them with averaged in-distribution activations; and interchange intervention, which swaps activations between clean and corrupted runs to identify key mediators \cite{wang2023ioi, zhang2024activationpatching}. 
Activation patching operates by replacing hidden activations between different forward passes to test their causal contribution to behavior. 
Patching can be performed \textit{online}, where interventions occur during the forward pass and influence subsequent computations, or \textit{offline}, where precomputed activations are substituted post hoc for analysis~\cite{heimersheim2024activationpatching}.

The activation patching technique typically involves two contexts: a \textit{source} (or clean) context denoted by $C$, and a \textit{target} (or corrupted) context denoted by $C'$. 
Patching activations from $C$ into $C'$ results in a \textit{manipulated} (or patched) context $C^*$, in which we aim to study the model’s behavior. 
We assess the effect of an intervention by comparing the model’s response before the intervention on $C'$ and after the intervention on the patched context $C^*$.  
Given a model $\mathbf{M}$ and its responses $r$ and $r'$ from contexts $C$ and $C'$, 
the logit difference, $\mathbf{M}[r|C^*] - \mathbf{M}[r'|C^*]$, is one of the most common metrics to assess the effect of an intervention~\cite{zhang2024activationpatching}.
In this study, we introduce the \textit{Causal Influence} (CI) score, inspired by the Indirect Effect score~\cite{stolfo2023cma}. 
Let $\tilde{r}$ denote the answer expected under a given hypothesis (not necessarily equal to $r$). The CI score is defined as:
\[
    \mathrm{CI} = \tfrac{1}{2}\Big[ \big(\mathrm{P}(\tilde{r}\,|\,C^*) - \mathrm{P}(\tilde{r}\,|\,C')\big)
    + \big(\mathrm{P}(r'\,|\,C') - \mathrm{P}(r'\,|\,C^*)\big) \Big].
\]
The first term quantifies how the intervention increases the probability of the expected answer $\tilde{r}$ relative to its baseline context $C'$, while the second term captures how it reduces the probability of the competing answer $r'$. 
A large positive CI value supports the hypothesis, whereas a near-zero score reflects chance-level influence.






\subsection{Experimental Setup}
\label{sec:exp_setup}


In each counting experiment, we present the model with a sequence of repeated items and query the number of occurrences.
The dataset includes two modalities (textual and visual) and several controlled variations.
In the textual modality, contexts consist of item lists followed or preceded by a counting question, forming two configurations: list-first and question-first.
Each list is either monotypic, containing repetitions of a single item type (e.g., ``apple, apple, apple''), or polytypic, containing multiple distinct items (e.g., ``apple, banana, orange'').
Questions are either specific, asking for the count of one object type (e.g., ``How many apples are in the list?''), or general, asking for the total count of all objects (e.g., ``How many fruits are in the list?'').
In the visual modality, synthetic images are generated with 1--9 objects of simple geometric shapes (e.g., square, circle) under similar monotypic and polytypic settings, paired with text prompts asking for the number of items.
Further details on the benchmarks, including full prompt templates and input examples, are provided in the Appendix~\ref{sec:app_taks_details}.

Our analysis is categorized into three main types: behavioral analysis, in which we evaluate the counting accuracy of various models in different scenarios; observational analysis, in which we passively analyze internal representations using PCA and cosine similarity; and causal mediation analysis, in which we study model behavior under causal intervention. 
Since the main focus of this study is the mechanistic interpretability of counting, we leave the behavioral analysis to Appendix~\ref{sec:app_behavioral}.
We conduct experiments on two well-studied open-source LLMs (Llama3~\cite{llama3} and Qwen2.5~\cite{qwen25}) and LVLMs (InternVL3.5~\cite{internvl35} and Qwen2.5-VL~\cite{qwen25vl}).
In the following sections, we present experiments on Qwen2.5 7B and Qwen2.5-VL 7B, with experiments on other models deferred to the appendix.
To ensure precise control of input variables, all evaluations are performed on a synthetic dataset designed to vary object identity and quantity in a systematic way. 
A minimal causal analysis on natural datasets is provided in Appendix~\ref{sec:app_natural}.

\section{Representational Emergence of Counting}
\label{sec:representational}

We analyze how numerical representations emerge across layers by examining hidden embeddings from both the prompt context and the generated answer tokens. To this end, we record the hidden embeddings of the generated numerical token for contexts containing one to nine repeated items. PCA is applied to all item embeddings, averaged across different item types, such as fruits in text or shapes in vision. The resulting projections reveal a consistent trend: representations of smaller numbers become separable in early layers, whereas larger numbers are distinguished only in deeper layers. This indicates a gradual encoding of numerical magnitude along the model hierarchy.

\begin{figure}[t]
    \centering
    \begin{subfigure}[b]{0.49\linewidth}
        \centering
        \includegraphics[width=\linewidth]{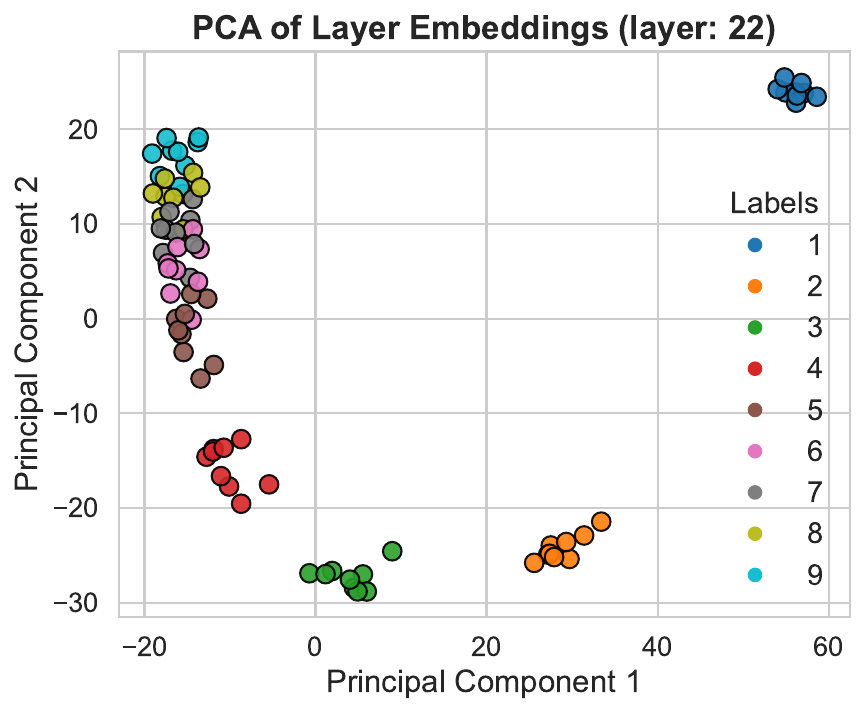}
    \end{subfigure}
    \begin{subfigure}[b]{0.49\linewidth}
        \centering
        \includegraphics[width=\linewidth]{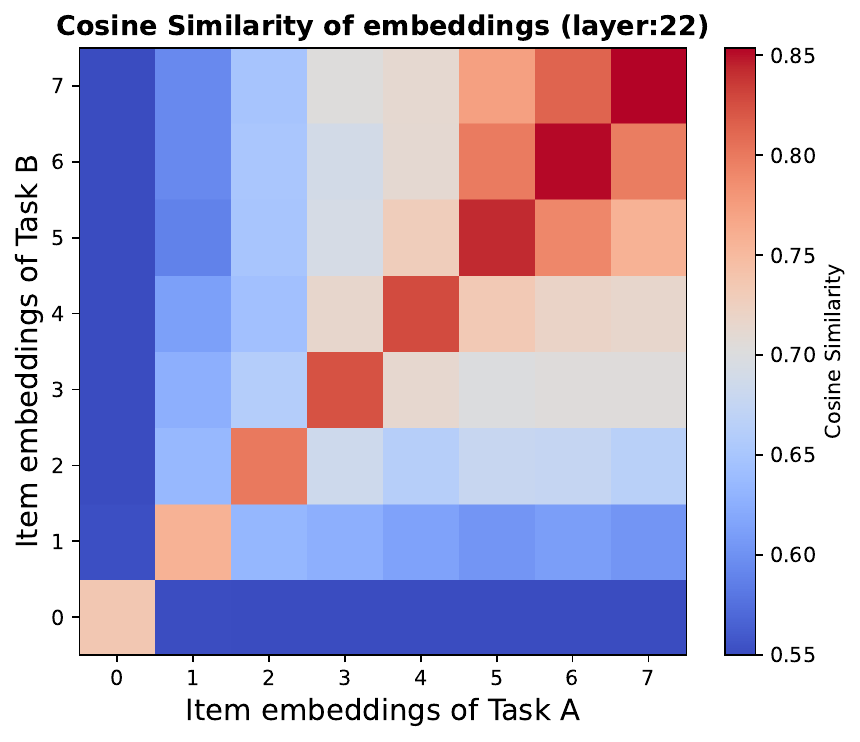}
    \end{subfigure}
    \vspace{-2pt}
    \caption{
    Representational behavior of embeddings in a selected layer of the LLM. 
    \textbf{Left:} PCA projection of embeddings corresponding to items at different list positions. 
    \textbf{Right:} Cross-task (different item types) cosine similarity of item embeddings, averaged over the dataset. 
    }
    \vspace{-8pt}
    \label{fig:representational}
\end{figure}

Figure~\ref{fig:representational} illustrates this pattern for layer~22, where PCA projections show clearly disentangled representations of items corresponding to different counts.
We also compute cosine similarity matrices between embeddings corresponding to list positions from different item types across layers, averaged over item types. These matrices exhibit a clear diagonal structure that emerges early for smaller counts and gradually extends toward higher counts across layers, as shown for layer~22 in Figure~\ref{fig:representational}. 
Another interesting observation is a logarithmic-like pattern~\cite{logarithmic}, where embeddings of higher counts are closer to each other in PCA space and more similar in cosine similarity.
Additional layerwise cosine similarity analysis and PCA visualizations of generated responses as well as separator tokens (e.g., commas) are provided in Appendix~\ref{sec:app_representational}. 
This suggests that LLMs and LVLMs develop structured internal encodings of count information based on token order and rhythmic patterns across tokens and layers.

\section{Causal Evidence of Counting Mechanism}

We next present causal analyses that identify where and how counting information is represented within the model activations. While the previous section revealed structured numerical patterns through observational analyses, causal interventions provide strong complementary evidence of counting mechanisms. Using activation patching and mediation-based methods, we assess which tokens, layers, and modalities carry information that determines the final numerical output. In the following subsections, we first introduce a specialized tool for counting analysis, then present coarse-grained and fine-grained localization results that reveal where numerical information resides within the model activations. Finally, we examine the underlying mechanisms that support counting across different configurations in both LLMs and LVLMs.
In this section, we report only aggregate metrics such as average probability and CI scores, averaged over all samples and counts. Detailed results and examples, as well as experiments on other models, are provided in Appendix~\ref{sec:app_causal}. 

\subsection{The CountScope Framework}

In our initial exploration, we used existing correlational tools like Logit Lens~\cite{nostalgebraist2020logitlens, neo2024llavainterp} to probe internal activations for count-based information. 
However, we found that these tools, while effective in decoding concrete concepts, were too noisy and unreliable when it came to numerical concepts. 
Even more advanced methods such as Tuned Lens~\cite{belrose2023tunedlens}, which aim to reduce mismatches between intermediate and final layers, fail to reveal meaningful patterns.  
This issue was especially pronounced in LVLMs, where the modality gap between image and text introduces additional complexity. 
To address these limitations, we developed a new causal tool, CountScope, specifically designed to probe numerical information with greater precision.

CountScope operates by creating a controlled, simple query setup that acts as a target context for isolating and revealing count-dependent information from various contexts and configurations. 
The core of CountScope is a context containing one or more placeholder tokens (textual or visual), followed by a counting question about that item. 
By applying minimal activation patching, similar to PatchScope~\cite{ghandeharioun2024patchscopes}, we transfer internal token activations from desired layers or tokens into the target context's placeholder token. 
The output predictions and softmax probabilities provide clear insights into what counting information is encoded within the selected tokens, layers, and modalities. 
This approach allows for a more granular understanding of how count-based information is represented within the model, and is applicable to both modalities, offering an accurate causal method for decoding internal representations.

\subsection{Input-Level Localization of Counting}
\label{sec:coarse_localization}

To understand how models perform counting, an essential question is where the count-dependent information resides within the input. In the coarse-grained causal analysis, each input can be divided into two components: the context, which contains either a list of items (for LLMs) or an image of objects (for LVLMs), and the question, which asks for the count. In LVLMs, the question always follows the image tokens. For consistency, we also place the question after the list in textual inputs. Because the context appears before the question, the decoder-only model does not yet know that the upcoming task involves counting. Therefore, the main hypothesis here is that the pivotal numerical information determining the final answer is stored either in the context or in the question segment.

Across all patching strategies, results consistently show that count-relevant information is primarily stored in the context rather than in the question. In the first experiment, offline zero patching was applied to the context tokens, setting their activations to zero. This caused a significant drop in the softmax probability of the ground truth number, with average values of $0.73 \pm 0.02$ for LLM and $0.65 \pm 0.31$ for LVLM. Offline zero patching of the question tokens had no significant effect, with a probability drop of $0.03 \pm 0.02$ for LLM and $0.05 \pm 0.03$ for LVLM.
The second experiment used offline interchange patching between two contexts containing different numbers of objects. The resulting predictions corresponded to the number from the source context, not the target, with a CI score of $0.61 \pm 0.02$ for LLM and $0.57 \pm 0.12$ for LVLM ($\tilde{r} = r$), confirming that numerical information is embedded in the contextual representation. Patching the question segments showed no significant effect, with a CI score of $0.02 \pm 0.01$ for LLM and $0.03 \pm 0.04$ for LVLM.

We next analyze which specific parts of the context carry this information. For textual tasks, offline zero patching of individual items revealed that removing the final item’s activation caused the largest reduction in ground-truth probability, with an average drop of $0.95 \pm 0.04$ for the correct number. This indicates that the final token encodes the ultimate count.
For vision-based tasks, we observe that a notable share of count-related information appears in background regions rather than in object patches. We assess this by transferring background or foreground patches into CountScope and recording the probability assigned to the correct number. Table~\ref{tab:BGvsFG} reports the average values for different image resolutions. Background patches generally carry stronger count signals, although this varies with image layout: when free space is limited, object patches become more informative.

\begin{table}[ht]
\centering
\resizebox{\columnwidth}{!}{
\begin{tabular}{lccc}
\hline
 & \textbf{3x3 Patches} & \textbf{6x6 Patches} & \textbf{10x10 Patches} \\ \hline
Foreground & $0.48_{\pm 0.02}$ & $0.46_{\pm 0.01}$ & $0.42_{\pm 0.01}$ \\ \hline
Background & $0.44_{\pm 0.03}$ & $0.58_{\pm 0.02}$ & $0.61_{\pm 0.01}$ \\ \hline
\end{tabular}
}
\vspace{-1pt}
\caption{Average CountScope probability of ground truth digits for foreground and background patches across different image sizes.}
\vspace{-5pt}
\label{tab:BGvsFG}
\end{table}

\begin{figure}[t]
    \centering

    \begin{subfigure}[t]{1\linewidth}
        \centering
        \includegraphics[width=\linewidth]{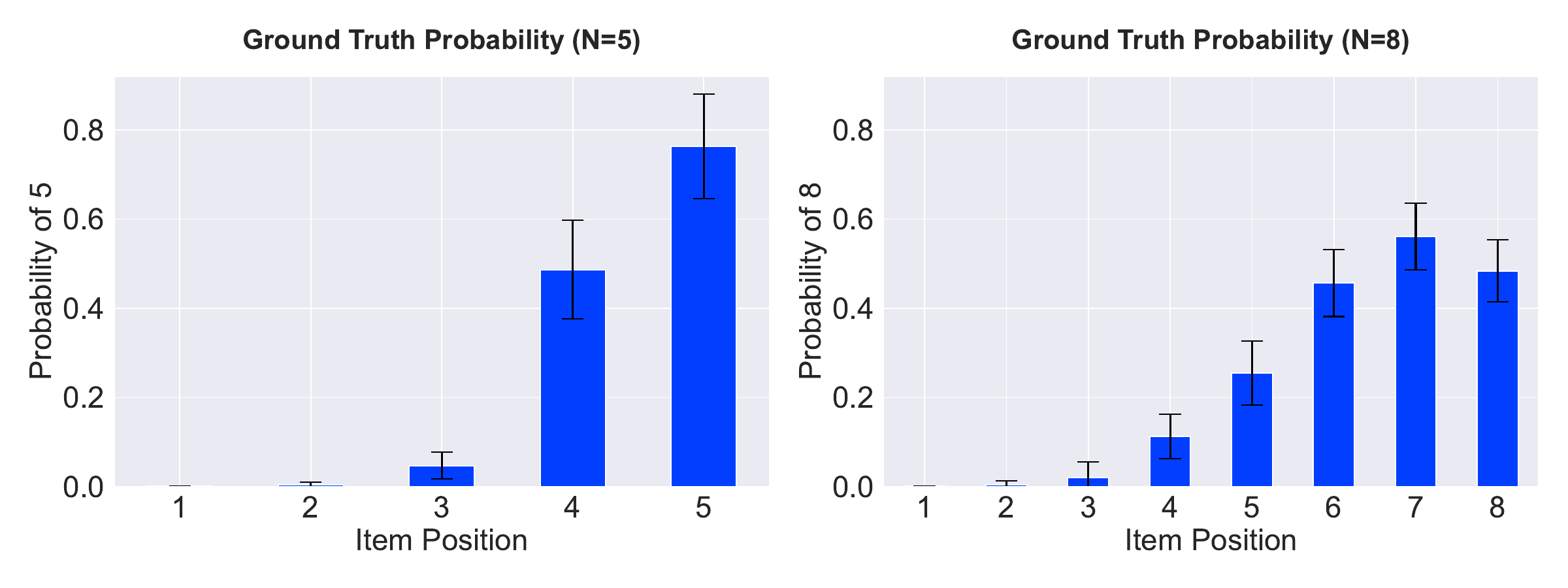}
    \end{subfigure}

    \vspace{-2pt}
    \caption{
        Ground-truth (total count) probability of visual objects decoded by CountScope.
        Besides the last object, nearby objects also contribute to the prediction,
        indicating distributed count information across adjacent regions. The order is based on positional embeddings and follows the causal attention.
    }
    \vspace{-8pt}
    \label{fig:item_information}
\end{figure}

For finer localization, we used visual CountScope and applied $3 \times 3$ patch-level interventions to the source visual context, covering the object token and one padding of the surrounding background. We then measured the ground-truth probability of selected regions using CountScope. Results (Figure~\ref{fig:item_information}) show that the last object region, typically located in the lower-right area of the image, most often carries the total count information. However, unlike textual tasks, the most informative visual region is not always associated with the final object. This difference likely stems from the vision encoder’s full spatial attention, which allows global integration before the decoding stage. Notably, LVLMs show reduced sensitivity beyond five objects, whereas textual models maintain their performance up to ten.

\subsection{Emergence of Internal Counters}
\label{sec:internal_counters}

A central question in understanding counting behavior is whether the model produces the final numerical response by summing information from all items simultaneously or by reading out a pre-computed count stored within the context. The causal analyses in the previous section show that the final number is encoded in the contextual representations, particularly in the final token or region. Building on this evidence, we now examine the internal mechanism that gives rise to this behavior. Specifically, through targeted interventions and controlled experiments, we test the hypothesis that LLMs and LVLMs maintain an \textbf{internal latent counter} that updates as the model processes each item.

\subsubsection{Per-Item Latent Count Encoding}

\begin{figure*}[t]
    \centering
    \begin{subfigure}[t]{0.32\linewidth}
        \centering
        \includegraphics[width=\linewidth]{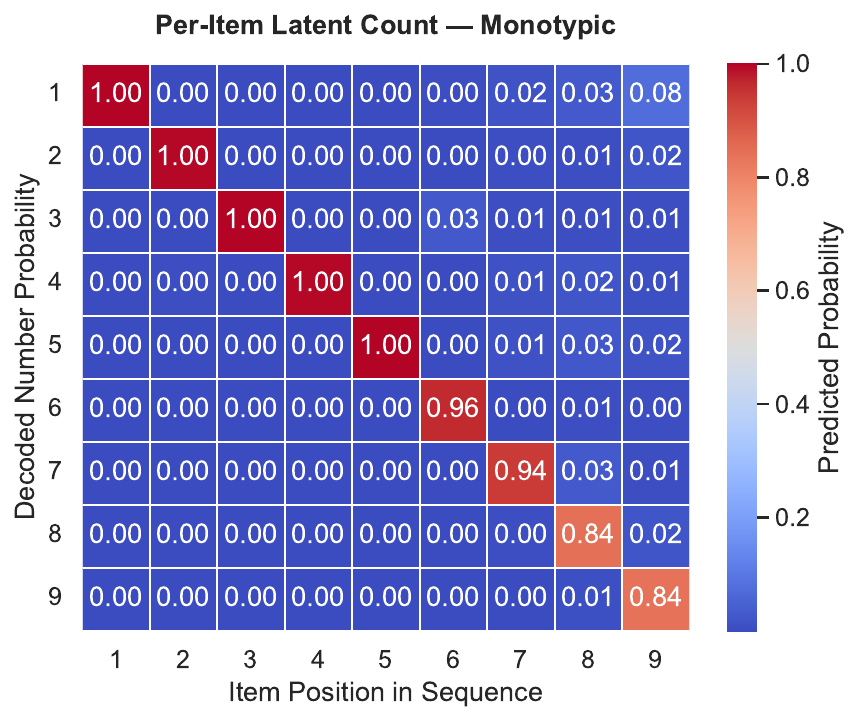}
        \vspace{-2pt}
        \caption{Textual (monotypic)}
    \end{subfigure}
    \hfill
    \begin{subfigure}[t]{0.32\linewidth}
        \centering
        \includegraphics[width=\linewidth]{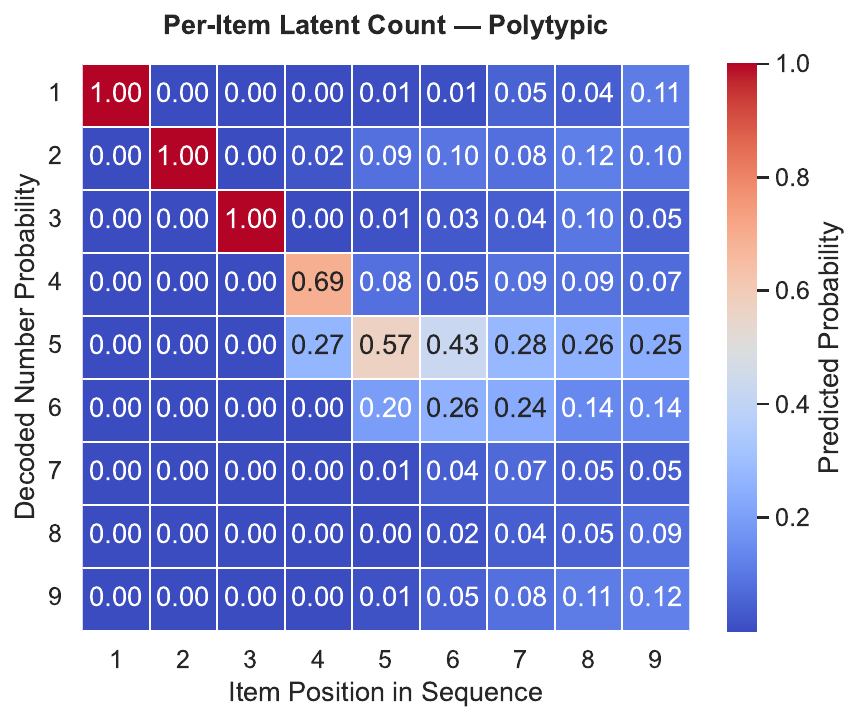}
        \vspace{-2pt}
        \caption{Textual (polytypic)}
    \end{subfigure}
    \hfill
    \begin{subfigure}[t]{0.32\linewidth}
        \centering
        \includegraphics[width=\linewidth]{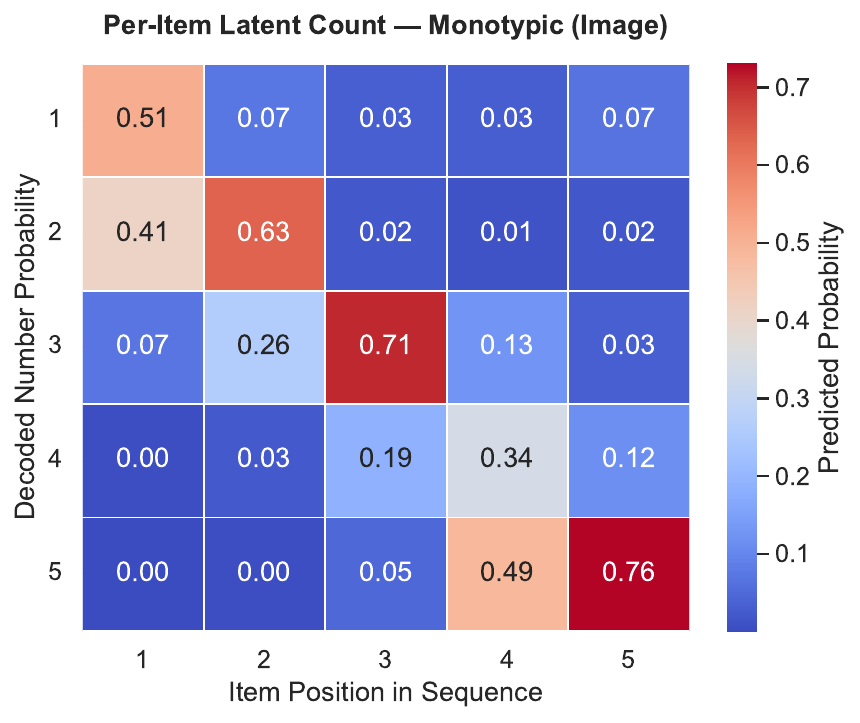}
        \vspace{-2pt}
        \caption{Visual (monotypic)}
    \end{subfigure}

    \vspace{-2pt}
    \caption{
        Per-item latent count encoding, decoded by CountScope.
        Each heatmap shows the average probability of decoding numbers (1–9) across sequence positions (1–9)
        for (a) textual monotypic, (b) textual polytypic with unique items, and (c) visual monotypic tasks.
        Textual tasks use the question-first configuration.
    }
    \vspace{-8pt}
    \label{fig:latentcount_peritem}
\end{figure*}

Our analyses from Section~\ref{sec:representational} suggest that the embeddings of successive items in a list encode progressively larger numerical values, forming a latent counting trajectory. To causally verify this, we probe each item using CountScope by transferring its activation into a minimal target context and recording the predicted counts and probabilities.
Figure~\ref{fig:latentcount_peritem} shows decoded probabilities for monotypic, polytypic, and visual tasks. Each item carries information about its sequence position, indicating that embeddings encode a latent representation of numerical order. The monotypic textual condition exhibits the clearest diagonal pattern, extending to larger counts, while the polytypic setting shows slightly weaker alignment due to object-type variability. 
In image-based tasks, a similar incremental pattern is observed, though less consistent across configurations. The clarity of this trend depends on factors such as the number of objects, their spatial arrangement, and visual distinctiveness within the image. Appendix~\ref{sec:app_natural} includes qualitative results on counting order across objects in natural images.

\subsubsection{Continued Counting}

To further investigate the internal counting mechanism, we design an interesting patching experiment. In this experiment, we patch the final $k$ items of the source context into the first $k$ items of the target context using online activation patching. Surprisingly, using CountScope, we observe that the $(k+1)$-th item in the target context does not carry the latent number $k+1$ (based on its actual position). Instead, it carries the latent number $N_{\text{source}} + 1$, where $N_{\text{source}}$ is the count of items in the source context. 
The final prediction of the target context is given by $\tilde{r} = N_{\text{source}} + N_{\text{target}} - k $. 
In other words, the internal counter behaves as if it simply continues from the end of the source list rather than restarting at the beginning of the target list.
Table~\ref{tab:continued_counting_types} shows the average CI score, providing strong support for this behavior.
Overall, these results indicate that the internal counter does not rely on recomputing a cumulative sum over all previous items in the list. Instead, it operates in a memoryless manner, updating based only on the most recent latent count rather than on the explicit number of all prior items.

\begin{table}[ht]
\centering
\resizebox{\columnwidth}{!}{
\begin{tabular}{lcccc}
\hline
\textbf{Model} & \textbf{Prompt Type} & \textbf{k=1} & \textbf{k=2} & \textbf{k=3} \\
\hline
\multirow{2}{*}{LLM}  & Question-First & $0.23_{\pm 0.09}$ & $0.9_{\pm 0.07}$ & $0.71_{\pm 0.17}$ \\
& Question-Last  & $0.10_{\pm 0.11}$ & $0.16_{\pm 0.05}$ & $0.16_{\pm 0.08}$ \\
\hline
\multirow{2}{*}{LVLM} 
& System-Prompt & $0.33_{\pm 0.07}$ & $0.43_{\pm 0.07}$ & $0.51_{\pm 0.07}$ \\
& User-Prompt   & $0.22_{\pm 0.05}$ & $0.42_{\pm 0.07}$ & $0.49_{\pm 0.07}$ \\
\hline
\end{tabular}
}
\vspace{-1pt}
\caption{Average CI scores for the continued counting hypothesis across different prompt structures for LLMs and LVLMs. As a question-first experiment for LVLMs, the counting task is defined in the system-prompt.
The hypothesis is most evident in the question-first setting for LLMs and in the system-prompt configuration for LVLMs.
}
\vspace{-5pt}
\label{tab:continued_counting_types}
\end{table}

\subsubsection{Type-Specific Resetting Counters}

We investigate the internal counting mechanism in polytypic settings, where the list contains diverse types of items, using CountScope for decoding. In this context, we observe that the confidence of the decodings is slightly lower, particularly in the list-first configuration. We focus on a specific type of item arrangement, where items belonging to the same type are grouped together, such as in the sequence: \texttt{apple, apple, orange, orange, orange, peach, peach, peach, peach.}

\begin{figure}[ht]
\centering
\includegraphics[width=\linewidth]{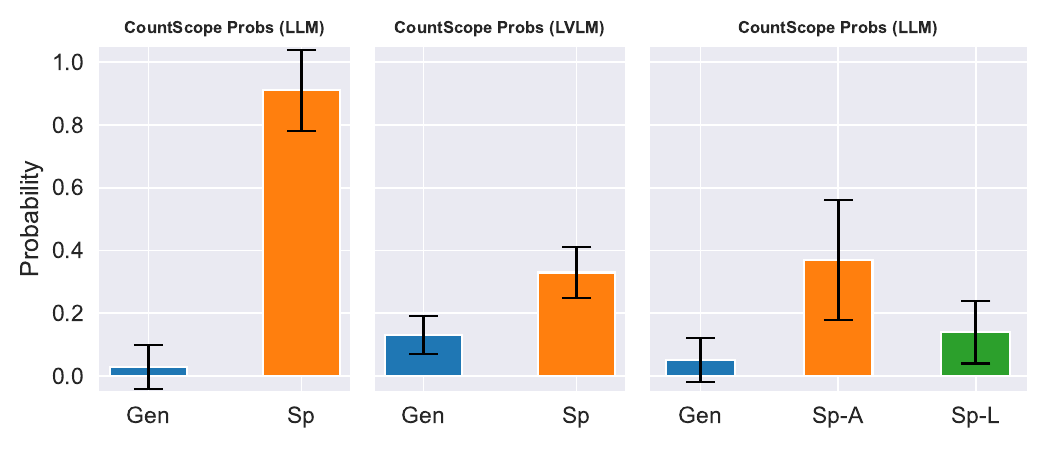}
\vspace{-2pt}
\caption{Type-specific counter behavior revealed by CountScope. Each bar shows the average probability of digits, averaged over 500 different configurations and 9 digits. Gen indicates total number of items, while Sp denotes the specific counter. Sp-A refers to the total number of separated items of a type, and Sp-L indicates the count of the last group of that type. Left: textual task; middle: visual task; right: textual task with repeated groups of one type.}
\vspace{-8pt}
\label{fig:counter:type_specific}
\end{figure}

In such cases, we do not observe a single global counter. Instead, the model maintains separate counters for each object type. For example, in the sequence above, the latent count for apples is 2, for oranges is 3, and for peaches is 4. When the same type reappears after an interruption by another type (e.g., \texttt{apple, apple, orange, orange, orange, apple, apple, apple, apple}), the counter for that type resets, producing a final latent count of 4 rather than 6. However, when the interruption is minimal (e.g., separated by only one different item), the counter continues without resetting, indicating a distance-dependent reset behavior.
Figure~\ref{fig:counter:type_specific} shows that for both LLMs and LVLMs the average probability of type-specific counts is significantly higher than that of the overall count. When items of one type are separated by only a short interval, the counter does not reset and continues accumulating.

\subsection{Influence of Maximum Latent Count}

In Section~\ref{sec:coarse_localization}, we observed that the final items of a list tend to carry the model's final numerical representation. Here, we extend this analysis by asking a related question: what happens if we swap the final item between two inputs that contain different numbers of elements? To address this question, we prepare two contexts, each containing a different number of repeated items, and interchange the activations of their final items. If the final item alone determines the numerical output, we would expect the model’s final prediction to always follow the count of the source context. However, this expectation does not hold consistently.

\begin{table}[ht]
\centering
\resizebox{\columnwidth}{!}{
\begin{tabular}{lccccc}
\hline
\multirow{2}{*}{\textbf{Model}} &
\multirow{2}{*}{\textbf{Prompt Type}} &
\multicolumn{2}{c}{\textbf{$N_{\text{source}} < N_{\text{target}}$}} &
\multicolumn{2}{c}{\textbf{$N_{\text{source}} > N_{\text{target}}$}} \\
\cline{3-6}
& & \textbf{k=1} & \textbf{k=2} & \textbf{k=1} & \textbf{k=2} \\
\hline
\multirow{2}{*}{LLM}  & Question-First & $0.25_{\pm 0.15}$ & $0.49_{\pm 0.16}$ & $0.54_{\pm 0.19}$ & $0.80_{\pm 0.13}$ \\
 & Question-Last  & $0.05_{\pm 0.02}$ & $0.10_{\pm 0.05}$ & $0.11_{\pm 0.06}$ & $0.32_{\pm 0.09}$ \\
\hline
\multirow{2}{*}{LVLM} & System-Prompt  & $0.76_{\pm 0.05}$ & $0.87_{\pm 0.02}$ & $0.82_{\pm 0.04}$ & $0.89_{\pm 0.03}$ \\
& User-Prompt    & $0.66_{\pm 0.05}$ & $0.80_{\pm 0.02}$ & $0.66_{\pm 0.06}$ & $0.87_{\pm 0.02}$ \\
\hline
\end{tabular}
}
\vspace{-1pt}
\caption{
Average CI scores for the maximum latent count hypothesis across different prompt types.
The hypothesis is more pronounced in the question-first (and system-prompt) setting.
}
\vspace{-7pt}
\label{tab:maximum_latent_count}
\end{table}

Empirically, when the source context contains more items than the target, the model’s final prediction aligns with the source count ($\tilde{r} = N_{\text{source}}$). In contrast, when the source context has fewer items, the prediction tends to follow the target count reduced by one ($\tilde{r} = N_{\text{target}} - 1$).
When two final items (e.g., \texttt{apple, apple}) are patched instead of one, the same pattern holds: if the source has more items, the prediction equals $N_{\text{source}}$, while if it has fewer, it approximates $\max(N_{\text{target}} - k, N_{\text{source}})$. Table~\ref{tab:maximum_latent_count} summarizes the corresponding CI scores, which show statistically significant differences consistent with this hypothesis.
These results indicate that the model does not simply copy the final number from the last item for its final response. Instead, it appears to evaluate the entire context and retrieve the \emph{maximum latent count} within it.

\subsection{Layerwise Emergence of Counting}
\label{sec:layerwise}

\begin{figure}[t]
    \centering
    \begin{subfigure}[t]{0.74\linewidth}
        \centering
        \includegraphics[width=\linewidth]{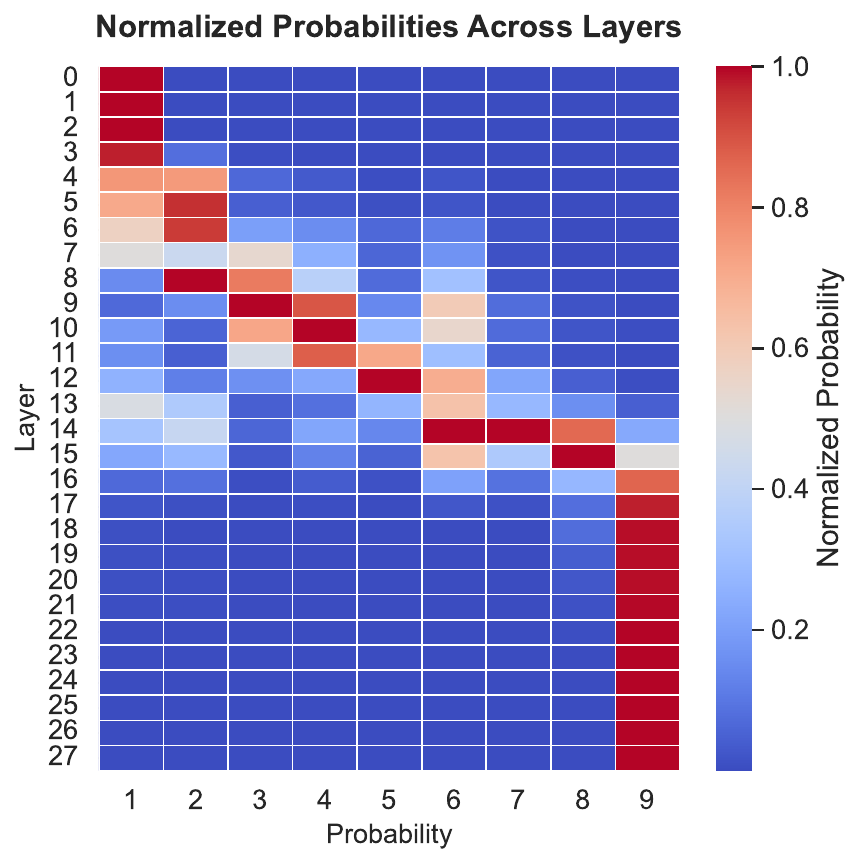}
        \caption{LLM}
    \end{subfigure}
    \begin{subfigure}[t]{0.74\linewidth}
        \centering
        \includegraphics[width=\linewidth]{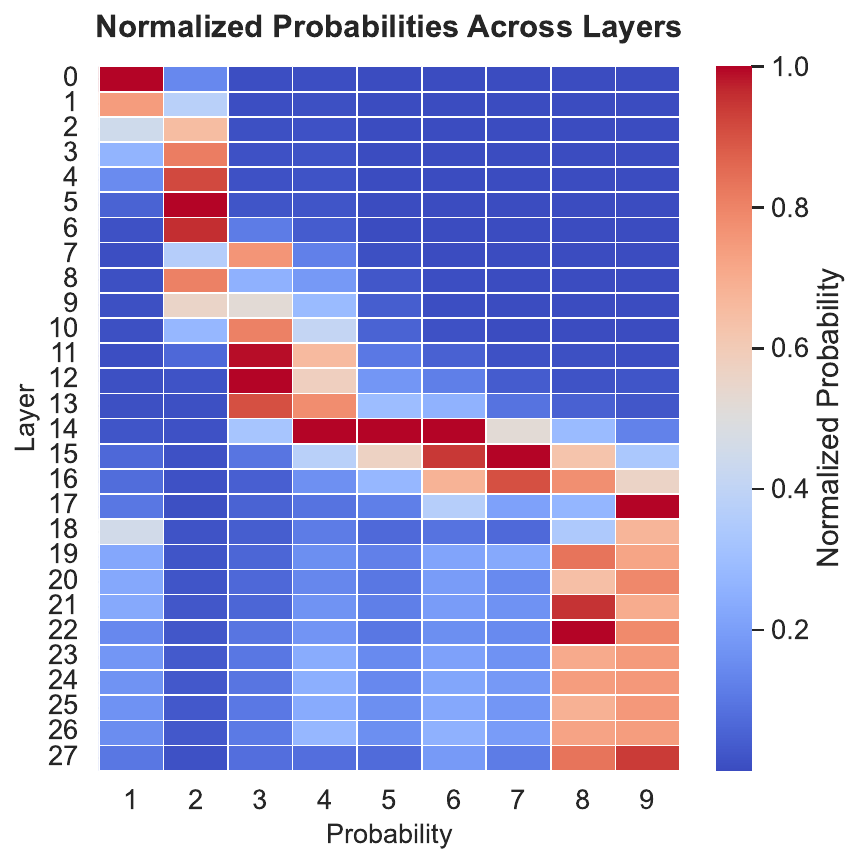}
        \caption{LVLM}
    \end{subfigure}
    \vspace{-2pt}
    \caption{
    Layerwise decoding of latent counts using online patching. CountScope is applied to layers 1 through $L$ of a context containing 9 items. Probabilities are normalized to the maximum value for each column.
    }
    \vspace{-8pt}
    \label{fig:latentcount_layerwise}
\end{figure}

We next examine the layerwise locations where latent count information is stored within the embeddings. To do this, we conduct two types of patching experiments. We apply CountScope, but instead of transferring all layers of the source item, we transfer embeddings only up to layer $L$ and generate the remaining embeddings through the forward pass. Figure~\ref{fig:latentcount_layerwise} shows how counts from 1 to 9 are progressively decoded from layers $L \approx 5$ to $L \approx 16$ in both LLM and LVLM. This demonstrates how count information emerges as the model deepens, providing insight into the layers where this information is encoded.
This result also reveals a key limitation of transformer architectures: higher layers are necessary to represent larger counts. Consequently, the model’s ability to count scales with layer depth, which may explain why LLMs have difficulty counting larger numbers of items.


\subsection{Linear Additivity}

We investigate whether latent number representations behave additively within the embedding space.
We conduct a mean intervention by computing the mean embedding for each list position, averaged across item types. We then form a position-difference vector for each layer by subtracting the mean embedding of position $i$ from that of position $j$.
This vector represents the change in latent count between the two positions.
To test linear additivity, we modify the embedding of a target item by adding the corresponding position-difference vector. For example, to shift an \texttt{apple} in the third list position so that it behaves as if it were in the sixth position, we add the position-difference vector for positions 3 and 6. This intervention is applied during the forward pass so that the adjusted embedding propagates through the remaining layers.

\begin{table}[ht]
\centering
\resizebox{\linewidth}{!}{
\begin{tabular}{lccccc}
\hline
\textbf{Model} & \textbf{k=1} & \textbf{k=2} & \textbf{k=3} & \textbf{k=4} \\ \hline 
LLM & $0.69_{\pm 0.05}$ & $0.66_{\pm 0.08}$ & $0.85_{\pm 0.11}$ & $0.60_{\pm 0.12}$ \\ \hline
LVLM & $0.47_{\pm 0.05}$ & $0.62_{\pm 0.07}$ & $0.33_{\pm 0.08}$ & $0.25_{\pm 0.07}$ \\ \hline
\end{tabular}
}
\caption{Average CI scores for the linear additivity hypothesis with position-difference of 1, 2, 3, and 4. Only layers 21 to 26 of both models are mediated.}
\label{tab:linear_additivity}
\end{table}

Table~\ref{tab:linear_additivity} shows the CI scores for the additivity test, indicating that the model’s predicted number shifts toward the intended value. The softmax probability of the target number increases and often becomes the highest. This pattern also holds when the mean vectors are estimated from a different counting task (see Appendix~\ref{sec:app_causal}), such as counting animals, demonstrating a clear disentanglement of latent numerical information in the representation.

\subsection{Shortcut Behavior on Separators}

\begin{figure}[t]
    \centering
    \begin{subfigure}[t]{0.76\linewidth}
        \centering
        \includegraphics[width=\linewidth]{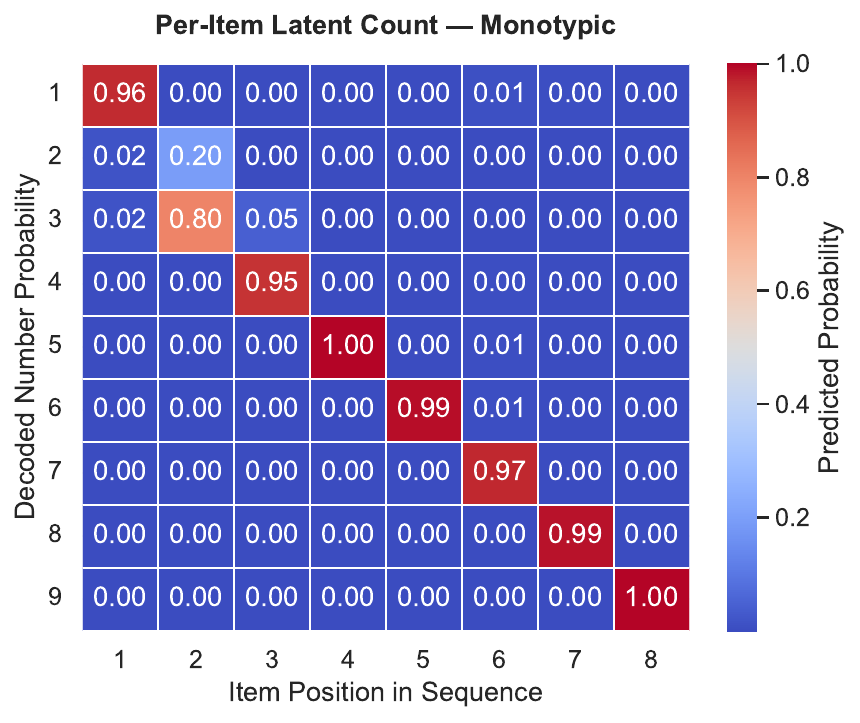}
        \caption{Monotypic}
    \end{subfigure}
    \begin{subfigure}[t]{0.76\linewidth}
        \centering
        \includegraphics[width=\linewidth]{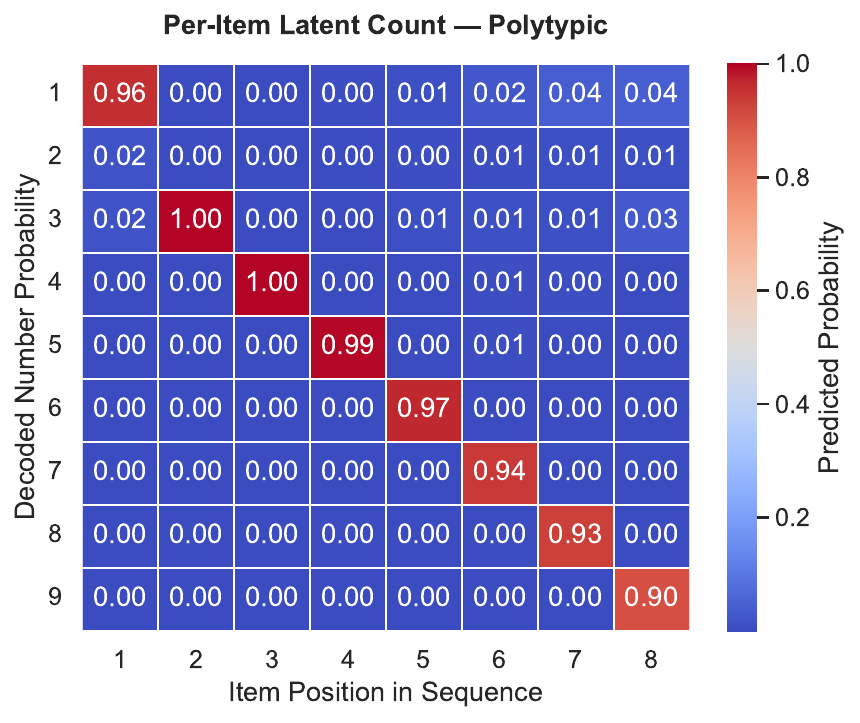}
        \caption{Polytypic}
        
    \end{subfigure}
    \vspace{-4pt}
    \caption{
    Per-item latent count for separators, decoded by CountScope.
    Each row shows the probability of decoding the count at different sequence positions for separators in (a) monotypic and (b) polytypic settings.
    }
    \vspace{-14pt}    
    \label{fig:latentcount_shortcut}
\end{figure}

Our analyses indicate that both element and separator tokens carry meaningful numerical information. To explore this further, we use CountScope to examine the embeddings of separators. Figure~\ref{fig:latentcount_shortcut} shows that separators alone contain sufficient information about their position in the list. Surprisingly, as compared to Figure~\ref{fig:latentcount_peritem}, we find that separators provide even more predictive signals than the elements, particularly in polytypic settings or in monotypic settings with higher numbers. Additionally, we observe that the first separator carries information about the first position, while subsequent separators encode information about the next position (e.g., the third comma reflects the number 4, not 3).

To further test the model’s reliance on separators over elements, we assess how the model’s final response changes when the information carried by separators conflicts with that of the elements. For this purpose, we patch all separator tokens in the context with the activation of the first separator in the list. Interestingly, we observe a significant performance drop of $0.75 \pm 0.39$ and $0.97 \pm 0.05$ for monotypic and polytypic settings, respectively, as measured against the ground truth number. This result indicates that the model’s output follows the pattern implied by the patched separators rather than the actual items, suggesting that the model heavily relies on separators as structural shortcuts for counting. Appendix~\ref{sec:app_behavioral} includes a behavioral analysis of counting failures due to disordered usage of separators.

\section{Practical Implication}

The main focus of this study is on the mechanistic interpretability of counting tasks. In this section, we provide an intuitive practical implication based on the analysis in Section~\ref{sec:layerwise}, which shows that counting is fundamentally limited by transformer depth. A promising direction is to use chain-of-thought (CoT)~\cite{COT} strategies that extend computation across tokens rather than relying only on model depth. Here, we present an orthogonal perspective on System-2-like strategies by investigating visual tokens as an additional capacity for numerical computation, instead of leveraging generated text tokens as in existing CoT-based methods~\cite{zhang2024tokenization, viser, grounding_id}.

\begin{figure}[t]
    \centering
    \begin{subfigure}[t]{0.78\linewidth}
        \centering
        \includegraphics[width=\linewidth]{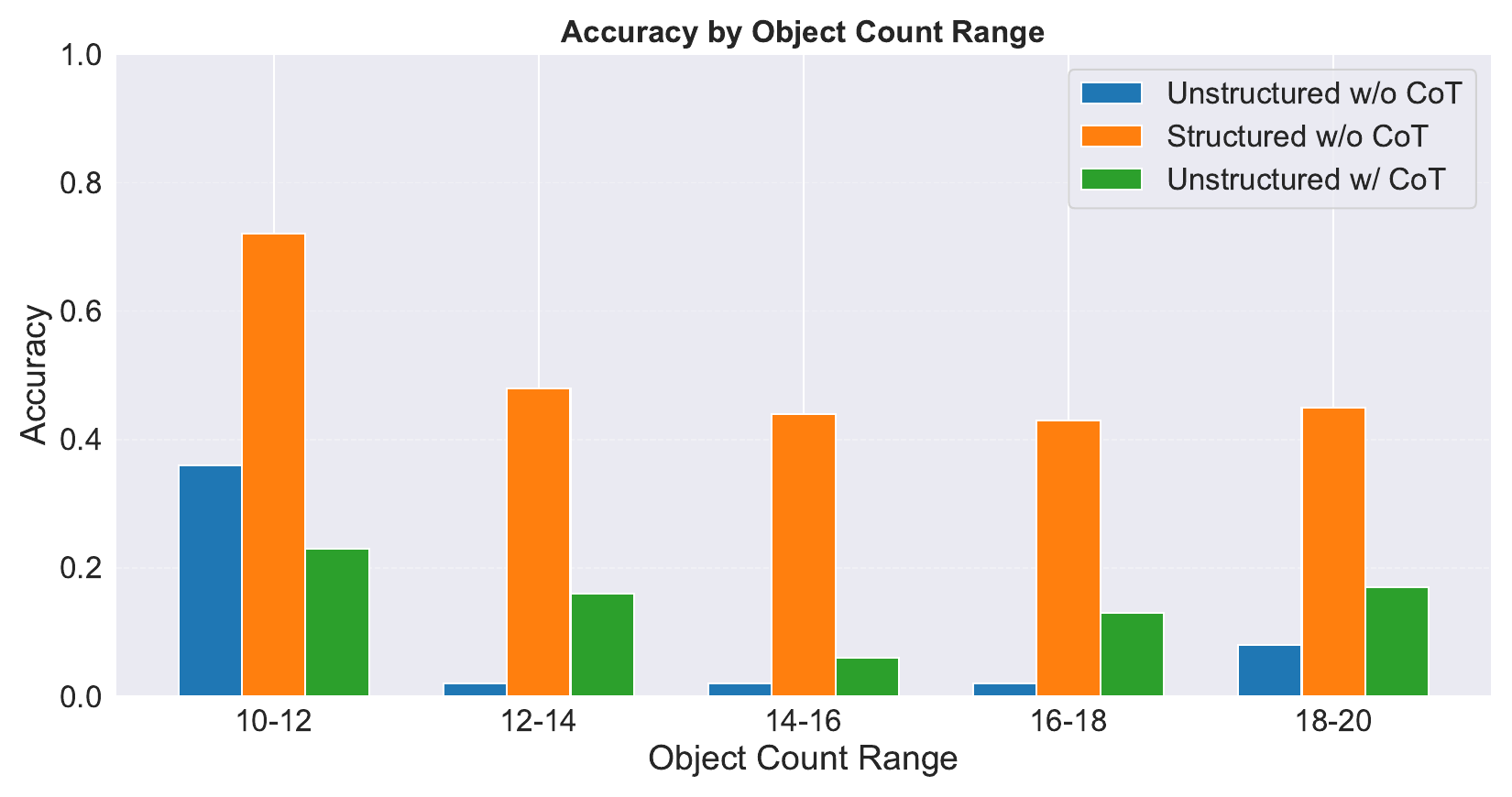}
        \caption{Behavioral results}
    \end{subfigure}
    \begin{subfigure}[t]{0.67\linewidth}
        \centering
        \includegraphics[width=\linewidth]{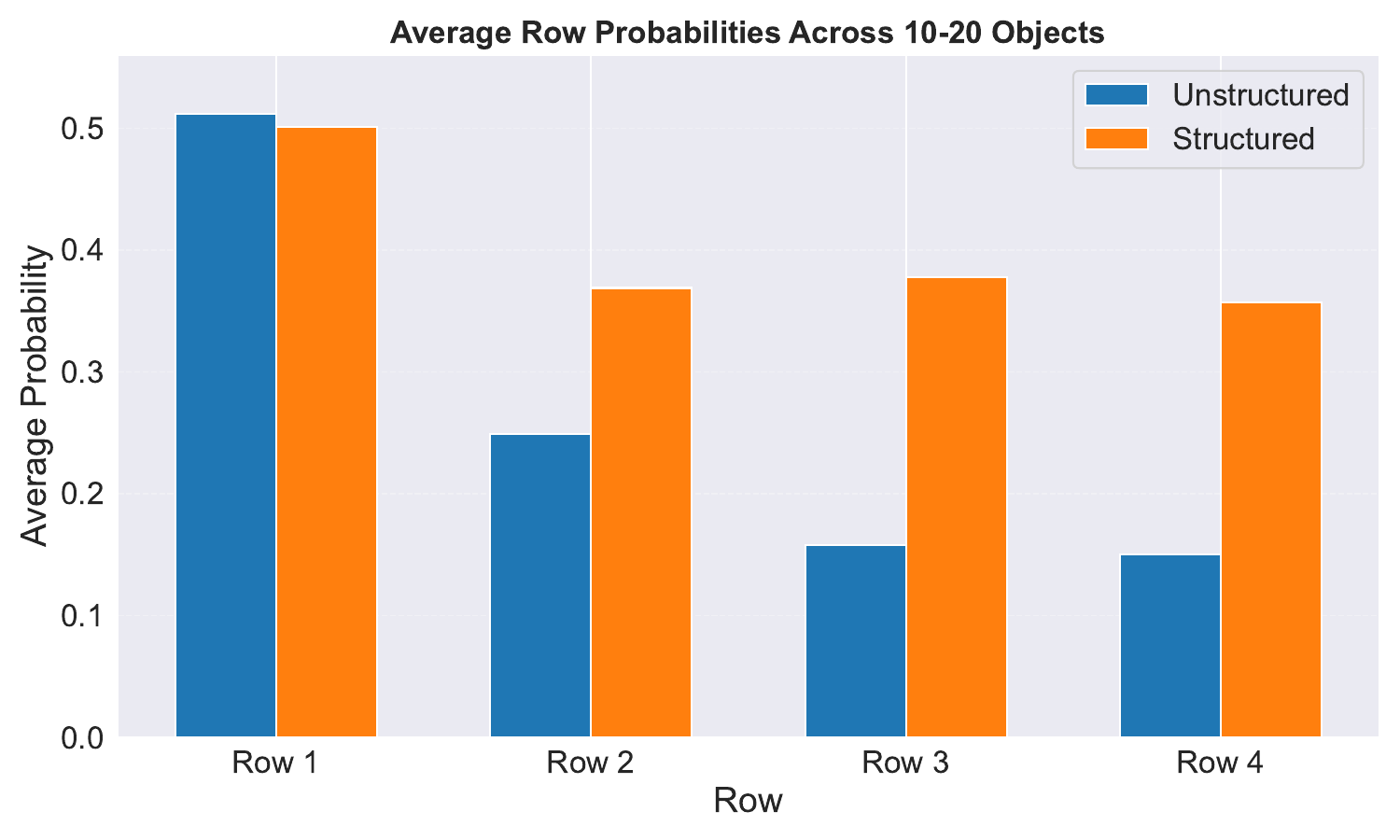}
        \caption{Causal analysis with CountScope}
    \end{subfigure}
    \vspace{-3pt}
    \caption{
    Effect of spatial structure on counting performance and internal representations. 
    \textit{Top:} Counting accuracy for 10--20 objects under three settings: unstructured image without CoT, structured without CoT, and unstructured with CoT. 
    \textit{Bottom:} Average CountScope probability assigned to the correct count for each row, averaged over 10--20 objects. 
    For the baseline without explicit partitioning, rows are defined virtually. Due to the causal attention mechanism, in the first row, both structured and unstructured settings show a similar decoding probability.
    }
    \vspace{-10pt}
    \label{fig:practical_implication}
\end{figure}

We partition images with lines into four regions to test whether imposed spatial structure provides additional capacity for numerical computation and storage. The model is prompted to output the total object count as a single digit, avoiding textual CoT to isolate the visual effect. The counting task is also mentioned in the system prompt, which precedes the visual tokens. Figure~\ref{fig:practical_implication} shows both behavioral and causal results of this experiment. Counting accuracy for 10--20 objects increases from 10.0\% in the baseline to 50.4\% with partitioning, while CoT without image partitions achieves an average accuracy of 15.3\%. Applying CountScope to each partition shows that the imposed structure encourages the model to decompose counting into smaller sub-problems. The average ground-truth probability of the last three partitions is 0.37 with structure, compared to 0.18 in the unstructured baseline, supporting the counter-resetting behavior observed across partitions.
\section{Findings and Discussion}

Our analyses reveal a coherent picture of how LLMs and LVLMs represent and compute numerical information. Despite differences between modalities, both model types exhibit similar internal mechanisms and patterns of numerical representations.
We introduce CountScope, a causal decoding tool that localizes count-relevant signals across layers and tokens in both text and vision settings. The same principle can be applied across a wide range of linguistic and visual tasks, making this idea broadly applicable for studying model internals.

In this work, we aim to provide a deep understanding of a simple but fundamental function in transformer-based language models.
In summary, our experiments show that counting arises as a structured, layerwise process rather than as a single, monolithic operation. Numerical information appears early and then refines progressively: lower layers reliably encode small counts while deeper layers are required to represent larger values. This layerwise emergence explains why models struggle to count long sequences of items. Based on this observation, we also introduce a practical implication that encourages LVLMs to better utilize the capacity across visual tokens. Our causal analysis confirms that most count-relevant signals are stored in the context and not in the question, with the final token carrying the strongest causal cue. In vision-language models, numerical cues are present in visual embeddings but are noisier; surprisingly, background regions frequently carry strong count signals, and visual layout, resolution, and object density all modulate performance.

Additional properties of the latent representations further clarify how counting is executed.
We identify a transferable internal counter: activations at specific tokens or regions encode a latent count that can be moved between contexts. This counter behaves in a Markov-like manner: when final items from a source are patched into a target, the counter continues from the source’s end rather than recomputing a full sum.
In polytypic lists, the model tends to maintain type-specific counters that often reset after interruptions, with reset behavior depending on the gap length. Separator tokens encode positional information and act as structural shortcuts: interfering with separators degrades counting far more than equivalent perturbations of the element tokens. The final reported number also reflects a context-level maximum latent count rather than a simple positional readout. Importantly, latent numerical states are approximately linearly additive: difference vectors between position-mean embeddings can be added to an item embedding to shift its decoded count toward a target position. This linearity supports simple interventions and shows that count information is at least partly disentangled in activation space.

Only a few studies have examined causal mechanistic interpretability in vision-language models, and our work adds to this small body of research. By analyzing counting across text and vision settings, we highlight both shared properties and clear limits of current architectures, including bounded capacity tied to depth and the reliance on structural shortcuts such as separators. These findings point to design targets for future models that need more reliable internal memory, stronger numerical abstraction, and better use of visual structure. Our work focuses on the core mechanism behind counting, but many open questions remain.
A complete circuit discovery is still missing, and counting under system-2 setups such as CoT decoding is not fully understood.

\clearpage

{
    \fontsize{8.9}{10.4}\selectfont
    \bibliographystyle{ieeenat_fullname}
    \bibliography{main}
}

\clearpage
\appendix
\maketitlesupplementary

\section{Related Work}
\label{sec:related work}

\paragraph{Counting competence and evaluation instability in LLMs.}
A growing body of evidence suggests that current LLMs exhibit \emph{fragile} counting ability even on simple tasks.
Fu et al.\ report systematic failures on character-counting queries (e.g., counting occurrences of a letter), showing that errors persist across models and stem from the intrinsic difficulty of repetition-tracking rather than token frequency or exposure~\cite{fu2024letters}.
Zhang et al.\ argue from computational and empirical perspectives that transformer LLMs lack an inherent mechanism for unbounded counting and that common subword tokenization schemes can further degrade performance by obscuring item boundaries~\cite{zhang2024tokenization}.
Complementing these findings, Ball et al.\ show that performance on deterministic tasks such as counting is highly sensitive to seemingly benign prompt/content variations, cautioning against extrapolating from single-prompt evaluations (the \emph{fixed-effect fallacy})~\cite{canwecount2024}.
Orthogonally, Yehudai et al.\ analyze \emph{when} transformers can count and show that exact counting of a token’s frequency in a string is feasible when the model state dimension scales linearly with context length, providing conditions and constructions that clarify capability limits~\cite{yehudai2024whencount}.
Together, these works motivate analyses that go beyond aggregate accuracy to \emph{how} numerical information is internally represented and why surface behavior can flip under minor rephrasings.

\vspace{-3mm}
\paragraph{Mechanistic and causal analyses of numeracy.}
A parallel line of work uses controlled tasks and interpretability tools to probe internal number mechanisms.
Golkar et al.\ introduce a contextual counting task and show that autoregressive transformers can learn effective counting strategies; intriguingly, removing positional embeddings or using rotary encodings changes where and how the mechanism emerges~\cite{golkar2024contextual}.
{Stolfo et al.\ apply a causal mediation analysis framework to arithmetic reasoning, intervening on layerwise activations to identify mid--late layers and specific components that causally mediate correct versus incorrect arithmetic predictions~\cite{stolfo2023cma}. In our work, we adopt compatible mediation-style \emph{effect scores} to quantify causal influence of activations on numeric outputs, in spirit similar to their CMA-based scoring.}
Shah et al.\ probe numeral embeddings and hidden states, finding cognitive-like magnitude effects (distance, ratio, size), hinting at an emergent ``mental number line''~\cite{shah2023numeric}.
Our work builds on these insights but focuses specifically on \emph{counting} and contributes a unified causal picture across both text-only LLMs and LVLMs: we show that latent count states are (i) primarily stored in contextual activations, (ii) concentrate in the final token/region, (iii) emerge progressively across layers, and (iv) are \emph{linearly additive} and transferable across contexts.

\vspace{-3mm}
\paragraph{Counting in LVLMs and the role of vision.}
In multimodal models, counting introduces additional challenges due to visual encoding, spatial aggregation, and modality fusion.
Paiss et al.\ demonstrate that CLIP---despite strong retrieval---often ignores cardinality; they improve counting via a counting-aware contrastive loss and show broader attentional coverage after fine-tuning~\cite{paiss2023clipcount}.
Guo et al.\ expose striking failures in \emph{compositional} counting (multiple object types), indicating unreliable binding between category and quantity~\cite{guo2025vlmcount}.
\emph{LLaVA-Interp} analyzes LLaVA’s \emph{visual} tokens with Logit Lens and ablations, showing that visual representations increasingly align with vocabulary space across layers and can project onto content-descriptive tokens (including numerals), suggesting image-derived numeric cues may surface in LM-space and relate to counting behavior~\cite{neo2024llavainterp}.
Consistent with these observations, we find that LVLMs’ count evidence often resides in both foreground and \emph{background} visual tokens and is more sensitive to layout, density, and resolution than in text-only models.
Our causal patching further shows that, unlike text, the most informative visual region is not always tied to the last object, reflecting pre-decoder global integration in the vision stack.

\vspace{-3mm}
\paragraph{Causal intervention toolkit and patch-based probes.}
Methodologically, our study connects to activation-level interpretability: activation/mean/interchange patching, key--value (attention) interventions, and mediation analysis; and to layerwise readout probes such as the original \emph{Logit Lens} and its learned variant \emph{Tuned Lens}~\cite{nostalgebraist2020logitlens,belrose2023tunedlens}.
\emph{Patchscopes} unifies patching/inspection configurations and leverages stronger models to explain internal representations in natural language, enabling expressive cross-layer analyses~\cite{ghandeharioun2024patchscopes}.
We introduce \textbf{CountScope}, a lightweight target-context probe tailored to numerical decoding, enabling precise, layerwise and token/patch-level localization of latent count states in both LLMs and LVLMs.
Beyond corroborating prior observations (e.g., layerwise emergence, shortcutting on separators), our causal experiments uncover (a) \emph{internal latent counters} that update across items and transfer across contexts, (b) \emph{type-specific counters} that can \emph{reset} in polytypic lists, and (c) a \emph{max-latent-count} effect whereby the model’s final prediction often reflects the maximum latent count present in context rather than a simple sum.

\section{Task Details}
\label{sec:app_taks_details}

The textual dataset is built from simple lists of item names and short counting questions. Items are sampled uniformly from a fixed vocabulary of common fruits (apple, orange, peach, fig, mango, pear, coconut, cherry, plum). Lists range from length 1 to 9. We use four prompt configurations: monotypic lists, polytypic lists, list-first (also called question-last) prompts, and question-first prompts. Both total-count and type-specific questions are included. The main templates used for our experiments are shown below.

\newtcolorbox{promptbox}[2][]{
  colback=gray!3,
  colframe=gray!50,
  coltitle=white,
  fonttitle=\bfseries,
  title={#2},
  boxsep=4pt,
  left=4pt,
  right=4pt,
  top=4pt,
  bottom=4pt,
  colbacktitle=black,
  #1
}
\begin{promptbox}{Monotypic • Question-first}
\texttt{Question: How many items are there in the following sentence?} \\
\texttt{apple, apple, apple, apple, apple}
\end{promptbox}

\begin{promptbox}{Polytypic • Question-first}
\texttt{Question: How many apples are there in the following sentence?} \\
\texttt{apple, peach, orange, pear, apple}
\end{promptbox}

\begin{promptbox}{Monotypic • Question-last}
\texttt{apple, apple, apple, apple, apple} \\
\texttt{Question: How many items are there in the above sentence?}
\end{promptbox}

\begin{promptbox}{Polytypic • Question-last}
\texttt{apple, peach, fig, apple, mango} \\
\texttt{Question: How many apples are there in the above sentence?}
\end{promptbox}

\begin{figure}[t]
    \centering
    \begin{subfigure}[t]{0.7\linewidth}
        \centering
        \includegraphics[width=\linewidth]{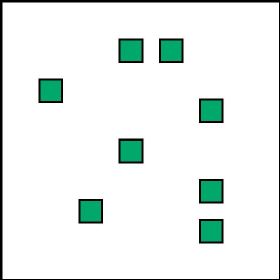}
        \caption{Monotypic}
    \end{subfigure}

    \begin{subfigure}[t]{0.7\linewidth}
        \centering
        \includegraphics[width=\linewidth]{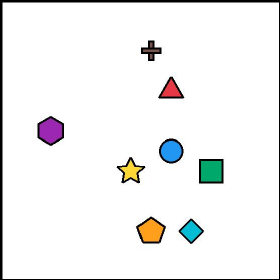}
        \caption{Polytypic-Unique}
    \end{subfigure}

    \caption{
    Example images from the visual dataset.  
    Monotypic images contain repeated instances of one shape–color pair, while polytypic images contain mixtures.  
    The prompt used for image queries is: \texttt{"How many objects are there in the image?"}}
    \label{fig:image_sample}
\end{figure}

The visual dataset consists of synthetic images containing one or more colored geometric shapes. Shapes are drawn from a fixed set (circle, triangle, square, pentagon, hexagon, star, diamond, cross, heart), and colors are drawn from (blue, green, red, yellow, orange, brown, purple and cyan). Each image contains between 1 and 9 shapes. Objects are placed at random non-overlapping positions on a uniform background using a simple sampling-and-rejection algorithm. Both monotypic and polytypic configurations are created by controlling shape–color combinations. Representative examples for each configuration are shown in Figure~\ref{fig:image_sample}. For causal analysis, we set the object size equal to the patch size of the model to avoid the confounding effect of multiple-patch aggregation. Since visual tokens always precede prompt tokens in LVLMs, the visual setup typically matches the question-last condition; however, by using a task description in the system prompt, we can approximate the question-first condition in visual experiments.

\section{Behavioral Characterization of Counting}
\label{sec:app_behavioral}

\begin{table}[t] 
\centering

\begingroup
\setlength{\tabcolsep}{1pt}  
\resizebox{\linewidth}{!}{
\begin{tabular}{@{}llcccc@{}}
\toprule
\textbf{Model} & \textbf{Category} &
\multicolumn{2}{c}{\textbf{Question-last}} &
\multicolumn{2}{c}{\textbf{Question-first}} \\
\cmidrule(lr){3-4}\cmidrule(lr){5-6}
 &  & Specific & General & Specific & General \\
\midrule
\multirow{3}{*}{\textbf{Qwen2.5}}
  & Monotypic          & 91.67 & 91.67 & 91.67 & 91.67 \\
  & Polytypic-Replicate & 88.13 & 91.96 & 88.34 & 70.90 \\
  & Polytypic-Unique   & 57.71 & 100.00 & 89.13 & 100.00 \\
\cmidrule(l){2-6}
\multirow{3}{*}{\textbf{Llama3}}
  & Monotypic          & 94.44 & 94.44 & 94.44 & 94.44 \\
  & Polytypic-Replicate & 94.35 & 6.23   & 89.85 & 1.41  \\
  & Polytypic-Unique   & 100.00 & 99.23 & 78.38 & 100.00 \\
\midrule
\multirow{3}{*}{\textbf{Qwen2.5-VL}}
  & Monotypic          & 100.00 & 100.00 & 100.00 & 100.00 \\
  & Polytypic-Replicate & 92.39 & 40.21  & 80.24 & 48.81 \\
  & Polytypic-Unique   & 100.00 & 100.00 & 100.00 & 100.00 \\
\cmidrule(l){2-6}
\multirow{3}{*}{\textbf{InternVL3.5}}
  & Monotypic          & 100.00 & 100.00 & 100.00 & 100.00 \\
  & Polytypic-Replicate & 99.36  & 56.05  & 99.77  & 64.58 \\
  & Polytypic-Unique   & 100.00 & 98.46  & 100.00 & 100.00 \\
\bottomrule
\end{tabular}
}
\caption{Average accuracy of LLMs and LVLMs on textual counting tasks across category types, ordering conditions, and question types. In the Polytypic-Replicate setting, some items are repeated (e.g., "apple, apple, orange"), while in the Polytypic-Unique setting, each item type appears only once (e.g., "apple, peach, orange").}
\label{tab:llm_behavioral}
\endgroup
\end{table}

We begin by quantifying the counting accuracy of LLMs and LVLMs across all experimental configurations. Table \ref{tab:llm_behavioral} reports the performance of two LLMs (Qwen2.5, Llama3) and two LVLMs (Qwen2.5-VL, InternVL3.5) on textual counting tasks across category types, ordering conditions, and question types. All models are of similar size (~7-8B parameters). Overall, LVLMs match or exceed LLM performance even though the tasks contain no visual input. Across all models, the polytypic-replicate setting is the most difficult, with clear drops in accuracy relative to monotypic and polytypic-unique. The polytypic-unique setting remains stable and often approaches monotypic accuracy. Accuracy differences between question-last and question-first formats are small. Specific questions show higher accuracy than general questions in polytypic conditions, which is mainly due to the smaller number of valid target types in the specific setting. Among all models, InternVL3.5 and Qwen2.5 show the strongest overall performance.

\begin{table}[t]
\centering
\setlength{\tabcolsep}{1pt} 
\resizebox{\columnwidth}{!}{
\begin{tabular}{@{}llcccc@{}} 
\toprule
\multirow{2}{*}{\textbf{Model}} & \multirow{2}{*}{\textbf{Category}} &
\multicolumn{2}{c}{\textbf{Question-Last}} &
\multicolumn{2}{c}{\textbf{Question-First}} \\
\cmidrule(lr){3-4}\cmidrule(lr){5-6}
 &  & Specific & General & Specific & General \\
\midrule
\multirow{3}{*}{\textbf{Qwen2.5-VL}}
  & Monotypic            & 30.56 & 41.67 & 41.67 & 44.44 \\
  & Polytypic-Replicate        & 76.75 & 62.26 & 73.72 & 65.48 \\
  & Polytypic-Unique        & 83.03 & 76.15 & 87.69 & 81.54 \\
\cmidrule(l){2-6}
\multirow{3}{*}{\textbf{InternVL3.5}}
  & Monotypic            & 94.44 & 91.67 & 69.44 & 55.56 \\
  & Polytypic-Replicate        & 86.89 & 81.07 & 71.70 & 75.19 \\
  & Polytypic-Unique       & 79.21 & 93.85 & 79.69 & 96.15 \\
\bottomrule
\end{tabular}
}
\caption{Accuracy of LVLMs on visual counting tasks under different category types, ordering conditions, and question types.}
\label{tab:lvlm_behavioral}
\end{table}

Table \ref{tab:lvlm_behavioral} summarizes LVLM behavior on visual counting tasks. Accuracy is lower than in textual tasks for the same numeric range. In contrast to the textual setting, both LVLMs show cases where polytypic inputs, including the replicate condition, outperform monotypic inputs, especially for Qwen2.5-VL. There is no consistent advantage for specific or general questions across models or ordering conditions. InternVL3.5 is stronger overall than Qwen2.5-VL in both question-first and question-last formats. For LVLMs, the question-first setup corresponds to placing the task description in the system prompt, which helps Qwen2.5-VL in several settings but does not produce a uniform pattern across models.


\begin{table}[t]
\centering
\setlength{\tabcolsep}{1pt}
\resizebox{\linewidth}{!}{
\begin{tabular}{@{}cccccccc@{}}
\toprule
\multirow{2}{*}{\textbf{Model}} &
\multirow{2}{*}{\textbf{Resolution}} &
\multicolumn{2}{c}{\textbf{Size 14}} &
\multicolumn{2}{c}{\textbf{Size 28}} &
\multicolumn{2}{c}{\textbf{Size 56}} \\
\cmidrule(lr){3-4}\cmidrule(lr){5-6}\cmidrule(lr){7-8}
 &  & Sparse & Dense & Sparse & Dense & Sparse & Dense \\
\midrule
\multirow{2}{*}{\textbf{InternVL3.5}}
 & 280$\times$280
   & 46.20 & 43.58
   & 67.22 & 64.38
   & 74.22 & 67.69 \\
 & 560$\times$560
   & 53.85 & 57.18
   & 73.50 & 74.51
   & 80.17 & 76.47 \\
\cmidrule(lr){2-8}
\multirow{2}{*}{\textbf{Qwen2.5-VL}}
 & 280$\times$280
   & 55.43 & 46.09
   & 70.28 & 74.52
   & 75.92 & 69.08 \\
 & 560$\times$560
   & 33.24 & 30.94
   & 66.10 & 63.87
   & 69.55 & 66.95 \\
\bottomrule
\end{tabular}
}
\caption{Accuracy by image resolution, object size, and density of objects for InternVL3.5 and Qwen2.5-VL.}
\label{tab:res_size_density_mixed}
\end{table}

Table \ref{tab:res_size_density_mixed} evaluates how image resolution, object size, and object density affect visual counting accuracy in InternVL3.5 and Qwen2.5-VL. The two models show opposite trends with respect to resolution: for Qwen2.5-VL, increasing the image size lowers accuracy, while InternVL3.5 benefits from higher resolution. This difference likely reflects model-specific optimal input resolutions. Both models improve substantially as object size increases, indicating that larger items provide clearer visual evidence for counting. Accuracy is consistently higher in the sparse setting than in the dense setting for all object sizes and both resolutions, suggesting that limited background area makes counting more difficult.

\begin{table}[t]
\centering
\resizebox{\linewidth}{!}{
\begin{tabular}{@{}llcccccccc@{}}
\toprule
\textbf{Model} & \textbf{Category} &
\multicolumn{4}{c}{\textbf{Question-last}} &
\multicolumn{4}{c}{\textbf{Question-first}} \\
\cmidrule(lr){3-6}\cmidrule(lr){7-10}
 &  & Various & Less & More & None & Various & Less & More & None \\
\midrule

\multirow{3}{*}{\textbf{Qwen2.5}} &
  Monotyp.               & 8.33 & 55.56 & 8.33 & 97.22 & 8.33 & 13.89 & 16.67 & 100 \\
 & Poly.-Rep.  & 5.53 & 8.29 & 11.95 & 27.43 & 6.32 & 8.04 & 16.75 & 31.61\\
 & Poly.-Uni.    & 17.69 & 22.30 & 76.57 & 75.38 & 32.30 & 27.69 & 61.21 & 76.15 \\

\midrule

\multirow{3}{*}{\textbf{Llama3}} &
  Monotyp.               & 16.67 & 8.33 & 55.56 & 19.44 & 11.11 & 11.11 & 44.44 & 25.00 \\
 & Poly.-Rep.  & 1.95 & 0.66 & 11.02 & 0.63 & 0.41 & 0.14 & 0.84 & 0.00 \\
 & Poly.-Uni.    & 40.00 & 14.62 & 76.15 & 30.77 & 43.08 & 3.85 & 93.85 & 6.15 \\

\bottomrule
\end{tabular}
}
\caption{Performance of Qwen2.5 and Llama3 under different separator conditions, including Various (some commas replaced with random separators), Less (some commas deleted), More (some commas repeated), and None (no separators).}
\label{tab:llm_separator}
\end{table}

We finally assess how different separator conditions impact counting accuracy in Table \ref{tab:llm_separator}. In all configurations, counting accuracy drops significantly when separators are altered or removed. This effect is observed in both question-last settings, where models must infer task instructions from context, and question-first settings, where the task is explicitly defined. The Polytypic-Replicate setting is particularly sensitive to changes in separators, likely due to the more complex composition of items. Interestingly, in the Monotypic setting, Qwen2.5 performs well even without separators (None condition), suggesting it can rely on the inherent structure of repeated items. In contrast, Llama3 shows much lower accuracy in the None condition, highlighting its greater reliance on separators, even when items are repetitive.

\section{Additional Causal Mediation Analysis}
\label{sec:app_causal}
\begin{table}[t]
\centering

\begingroup
\setlength{\tabcolsep}{4pt}
\resizebox{\linewidth}{!}{%
\begin{tabular}{@{}c|cc|cc@{}}
\toprule
 & \multicolumn{2}{c|}{\textbf{LLM (Textual)}} & \multicolumn{2}{c}{\textbf{LVLM (Visual)}} \\
\cmidrule(lr){2-3} \cmidrule(l){4-5}
\textbf{Count} &
\textbf{Context masked} &
\textbf{Question masked} &
\textbf{Image masked} &
\textbf{Prompt masked} \\
\midrule
1 & 0.0000 & 0.0000 & 0.9539 & 0.0187 \\
2 & 0.0654 & 0.0002 & 0.9691 & 0.0167 \\
3 & 0.6832 & 0.0000 & 0.9535 & 0.0304 \\
4 & 0.9806 & 0.0000 & 0.8789 & 0.0617 \\
5 & 0.5606 & 0.0000 & 0.7605 & 0.0714 \\
6 & 0.8135 & 0.0000 & 0.6363 & 0.1119 \\
7 & 0.7917 & 0.0000 & 0.5181 & 0.0617 \\
8 & 0.9497 & 0.2563 & 0.4916 & 0.0812 \\
9 & 0.9862 & 0.0008 & 0.4515 & 0.0330 \\
\bottomrule
\end{tabular}%
}
\endgroup

\caption{Mean drop in the probability of the ground-truth count under offline zero patching. For each count, we report the drop when masking text-only context vs.\ question tokens in the LLM, and image patches vs.\ prompt tokens in the LVLM. Higher values indicate that masking removes more count-relevant information from that part of the input.}
\label{tab:offline_zero_llm_lvlm}

\end{table}

\begin{figure}[t]
    \centering
    \includegraphics[width=\linewidth]{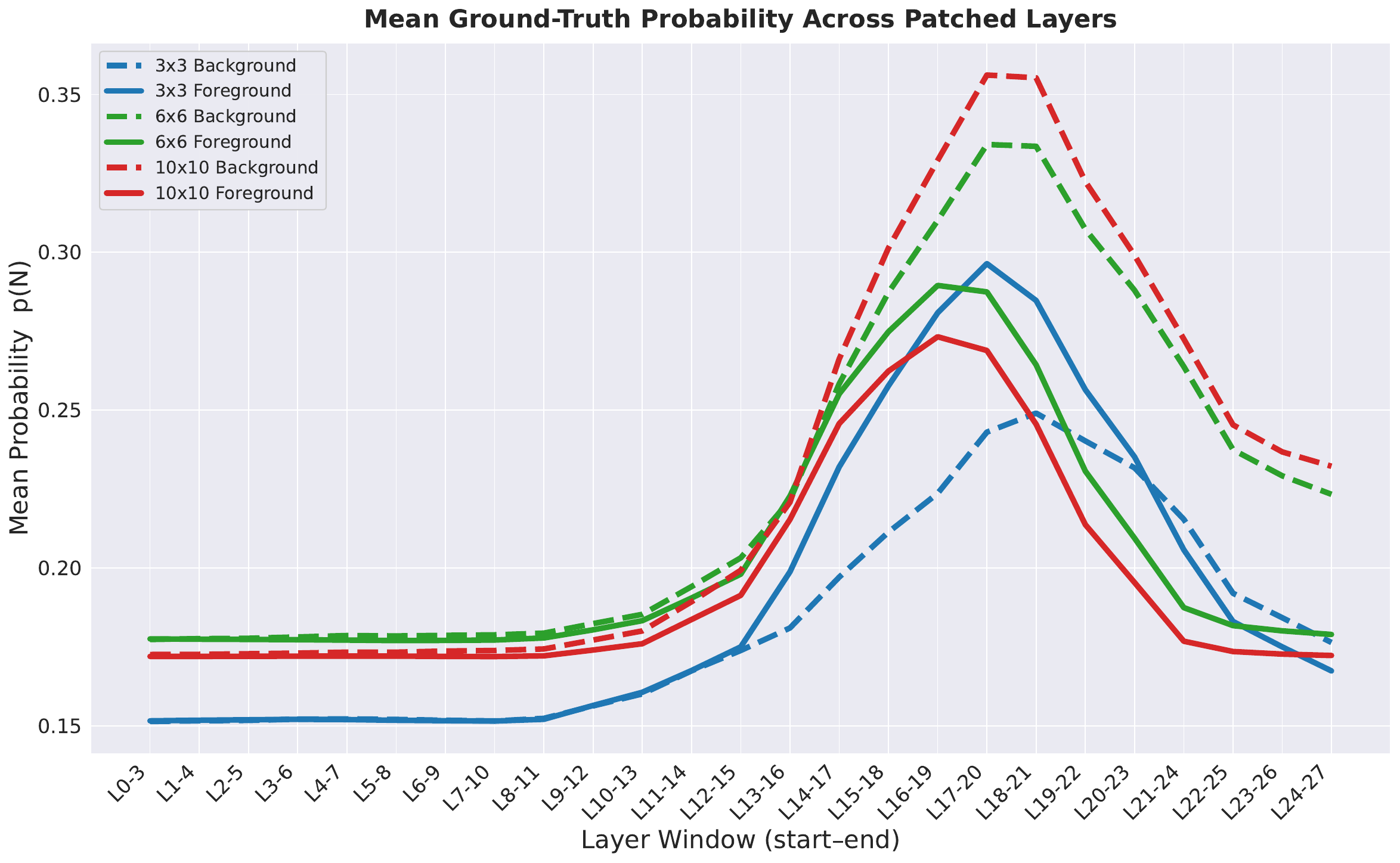}
    \caption{
    Mean ground-truth probability across layer windows under offline patch interchange.
    At each step, a window of four consecutive layers was interchanged, and the resulting
    ground-truth probability was measured. The reported values are averaged over images
    containing one to five objects. Each color corresponds to a specific image size
    (3$\times$3, 6$\times$6, and 10$\times$10), while solid and dashed lines distinguish
    foreground and background interchange, respectively.
    }
    \label{fig:offline_patch_interchange_layers}
\end{figure}

Here, we provide additional details of the experiments conducted for causal mediation analysis. Table \ref{tab:offline_zero_llm_lvlm} reports the mean drop in the probability of the ground-truth count after offline zero patching of context and question (for LLMs) or image and prompt (for LVLMs). The results confirm that count-related information is primarily stored in the context for both model types. Table \ref{tab:BGvsFG_per_object} provides a per-object breakdown of CountScope probabilities, revealing how the strength of count signals in foreground and background patches varies with the number of objects and patch size. Finally, Figure \ref{fig:offline_patch_interchange_layers} illustrates how the count information is distributed across layers under offline patch interchange, with background patches showing stronger signals in the early layers, and foreground patches becoming more informative in deeper layers. The diagrams also reveal that count information is mainly stored in the middle-to-late layers.

\begin{table}[t]
\centering

\begingroup
\setlength{\tabcolsep}{4pt}
\resizebox{\linewidth}{!}{%
\begin{tabular}{@{}cc|ccc@{}}
\toprule
\textbf{\#Objects} & \textbf{Region} &
\textbf{3x3 Patches} & \textbf{6x6 Patches} & \textbf{10x10 Patches} \\ 
\midrule

1 & Foreground & 0.4333 & 0.4392 & 0.4715 \\
1 & Background & 0.4980 & 0.5828 & 0.6255 \\
\midrule

2 & Foreground & 0.5827 & 0.5670 & 0.5481 \\
2 & Background & 0.5864 & 0.7141 & 0.7171 \\
\midrule

3 & Foreground & 0.5815 & 0.5715 & 0.4979 \\
3 & Background & 0.4826 & 0.6746 & 0.6892 \\
\midrule

4 & Foreground & 0.4629 & 0.3937 & 0.3219 \\
4 & Background & 0.3900 & 0.5568 & 0.5923 \\
\midrule

5 & Foreground & 0.3865 & 0.3516 & 0.2993 \\
5 & Background & 0.3343 & 0.4744 & 0.4899 \\
\bottomrule

\end{tabular}%
}
\endgroup

\caption{Per-object breakdown (1--5 objects) of the average CountScope probability for foreground and background patches across different patch sizes.}
\label{tab:BGvsFG_per_object}

\end{table}

\begin{table}[ht]
\centering

\begin{subtable}[t]{0.3\textwidth}
\centering
\resizebox{\columnwidth}{!}{
\begin{tabular}{ccccccccc}
\hline
\textbf{src / trgt} & \textbf{2} & \textbf{3} & \textbf{4} & \textbf{5} & 
\textbf{6} & \textbf{7} & \textbf{8} & \textbf{9} \\
\hline
\textbf{2}  & 3 & 4 & 5 & 6 & 7 & 8 & 9 & 10 \\
\textbf{3}  & 4 & 5 & 6 & 7 & 8 & 9 & 10 & 11 \\
\textbf{4}  & 5 & 6 & 7 & 8 & 9 & 10 & 11 & 12 \\
\textbf{5}  & 6 & 7 & 8 & 9 & 10 & 11 & 12 & 13 \\
\textbf{6}  & 7 & 8 & 9 & 10 & 11 & 12 & 13 & 14 \\
\textbf{7}  & 8 & 9 & 10 & 11 & 12 & 13 & 14 & 15 \\
\textbf{8}  & 9 & 10 & 11 & 12 & 13 & 14 & 15 & 16 \\
\textbf{9}  & 10 & 11 & 12 & 13 & 14 & 15 & 16 & 17 \\
\hline
\end{tabular}
}
\subcaption{$k=1$}
\end{subtable}
\hfill
\begin{subtable}[t]{0.3\textwidth}
\centering
\resizebox{\columnwidth}{!}{
\begin{tabular}{ccccccccc}
\hline
\textbf{src / trgt} & \textbf{2} & \textbf{3} & \textbf{4} & \textbf{5} & 
\textbf{6} & \textbf{7} & \textbf{8} & \textbf{9} \\
\hline
\textbf{2}  & 2 & 3 & 4 & 5 & 6 & 7 & 8 & 9 \\
\textbf{3}  & 3 & 4 & 5 & 6 & 7 & 8 & 9 & 10 \\
\textbf{4}  & 4 & 5 & 6 & 7 & 8 & 9 & 10 & 11 \\
\textbf{5}  & 5 & 6 & 7 & 8 & 9 & 10 & 11 & 12 \\
\textbf{6}  & 6 & 7 & 8 & 9 & 10 & 11 & 12 & 13 \\
\textbf{7}  & 7 & 8 & 9 & 10 & 11 & 12 & 13 & 14 \\
\textbf{8}  & 8 & 9 & 10 & 11 & 12 & 13 & 14 & 15 \\
\textbf{9}  & 9 & 10 & 11 & 12 & 13 & 14 & 15 & 16 \\
\hline
\end{tabular}
}
\subcaption{$k=2$}
\end{subtable}
\hfill
\begin{subtable}[t]{0.3\textwidth}
\centering
\resizebox{\columnwidth}{!}{
\begin{tabular}{ccccccccc}
\hline
\textbf{src / trgt} & \textbf{2} & \textbf{3} & \textbf{4} & \textbf{5} & 
\textbf{6} & \textbf{7} & \textbf{8} & \textbf{9} \\
\hline
\textbf{2}  & 1 & 2 & 3 & 4 & 5 & 6 & 7 & 8 \\
\textbf{3}  & 2 & 3 & 4 & 5 & 6 & 7 & 8 & 9 \\
\textbf{4}  & 3 & 4 & 5 & 6 & 7 & 8 & 9 & 10 \\
\textbf{5}  & 4 & 5 & 6 & 7 & 8 & 9 & 10 & 11 \\
\textbf{6}  & 5 & 6 & 7 & 8 & 9 & 10 & 11 & 12 \\
\textbf{7}  & 6 & 7 & 8 & 9 & 10 & 11 & 12 & 13 \\
\textbf{8}  & 7 & 8 & 9 & 10 & 11 & 12 & 13 & 14 \\
\textbf{9}  & 8 & 9 & 10 & 11 & 12 & 13 & 14 & 15 \\
\hline
\end{tabular}
}
\subcaption{$k=3$}
\end{subtable}

\caption{Expected answer, $\tilde{r}$, under the continued counting hypothesis.
The value is computed as 
$\tilde{r} = N_{\text{source}} + N_{\text{target}} - k$,
where $N_{\text{source}}$ and $N_{\text{target}}$ denote the number of objects in the source and target inputs, respectively, and $k$ is the number of patched items.}
\label{tab:last2first:expected}
\end{table}

The \textit{continued counting} hypothesis is evaluated across both visual and textual tasks. Table \ref{tab:last2first:expected} presents the expected predictions based on this hypothesis for various values of $k$. Table \ref{tab:last2first:visual} reports the performance of visual models (InternVL3.5 and Qwen2.5-VL) for different values of $k$ in both "Question First" and "Question Last" setups. For the textual tasks, the results are shown in Tables \ref{tab:l2fllamaQwenQf} and \ref{tab:l2fllamaQwenQl}, where Llama3 and Qwen2.5 are evaluated with different patch types (both, separators, and elements) under "Question-First" and "Question-Last" conditions. These tables report the average probability of the predicted answer ($\tilde{r}$), the average probability of the incorrect answer ($r^\prime$), and the CI score for each configuration.

\begin{table}[t]
\centering

\begingroup
\setlength{\tabcolsep}{2pt}
\resizebox{0.9\columnwidth}{!}{
\begin{tabular}{@{}llcccccc@{}}
\toprule
 & & \multicolumn{3}{c}{\textbf{InternVL3.5}} & \multicolumn{3}{c}{\textbf{Qwen2.5-VL}} \\
\cmidrule(lr){3-5} \cmidrule(lr){6-8}
\textbf{Question Type} & \textbf{K} &
$\mathbf{P}(\tilde{r})$ & $\mathbf{P}(r^\prime)$ & \textbf{CI} &
$\mathbf{P}(\tilde{r})$ & $\mathbf{P}(r^\prime)$ & \textbf{CI} \\
\midrule
\multirow{3}{*}{\textbf{Question-First}}
  & 1 & 0.25 & 0.42 & 0.24 & 0.32 & 0.46 & 0.32 \\
  & 2 & 0.39 & 0.15 & 0.43 & 0.48 & 0.40 & 0.42 \\
  & 3 & 0.45 & 0.08 & 0.50 & 0.53 & 0.27 & 0.49 \\
\cmidrule(l){2-8}
\multirow{3}{*}{\textbf{Question-Last}}
  & 1 & 0.36 & 0.54 & 0.29 & 0.24 & 0.56 & 0.23 \\
  & 2 & 0.56 & 0.16 & 0.57 & 0.49 & 0.38 & 0.43 \\
  & 3 & 0.53 & 0.11 & 0.59 & 0.54 & 0.27 & 0.49 \\
\bottomrule
\end{tabular}
}
\caption{ Average probabilities $\Pr(\tilde{r})$, $\Pr(r')$, and CI scores for the continued-counting hypothesis across different values of $K$ and prompt structures in LVLMs. In the question-first setting, the counting task is specified in the system prompt.}
\label{tab:last2first:visual}
\endgroup
\end{table}

\begin{table}[t] 
\centering

\begingroup
\setlength{\tabcolsep}{2pt} 

\resizebox{0.9\columnwidth}{!}{
\begin{tabular}{@{}llcccccc@{}}
\toprule
 & & \multicolumn{3}{c}{\textbf{Llama3}} & \multicolumn{3}{c}{\textbf{Qwen2.5}} \\
\cmidrule(lr){3-5} \cmidrule(lr){6-8}
\textbf{K} & \textbf{Patch Type} &
$\mathbf{P}(\tilde{r})$ & $\mathbf{P}({r^\prime})$ & \textbf{CI} &
$\mathbf{P}(\tilde{r})$ & $\mathbf{P}({r^\prime})$ & \textbf{CI} \\
\midrule
\multirow{3}{*}{\textbf{1}}
  & Both       & 0.03 & 0.26 & 0.33 & 0.00 & 0.54 & 0.23 \\
  & Separators & 0.01 & 0.88 & 0.00 & 0.00 & 1.00 & 0.00 \\
  & Elements   & 0.03 & 0.26 & 0.33 & 0.00 & 0.54 & 0.23 \\
\cmidrule(l){2-8}
\multirow{3}{*}{\textbf{2}}
  & Both       & 0.70 & 0.01 & 0.78 & 0.82 & 0.00 & 0.91 \\
  & Separators & 0.41 & 0.37 & 0.45 & 0.23 & 0.74 & 0.24 \\
  & Elements   & 0.17 & 0.19 & 0.42 & 0.58 & 0.02 & 0.78 \\
\cmidrule(l){2-8}
\multirow{3}{*}{\textbf{3}}
  & Both       & 0.86 & 0.19 & 0.67 & 1.00 & 0.28 & 0.72 \\
  & Separators & 0.69 & 0.26 & 0.55 & 0.57 & 0.69 & 0.30 \\
  & Elements   & 0.35 & 0.30 & 0.36 & 0.81 & 0.29 & 0.62 \\
\bottomrule
\end{tabular}
}
\caption{Average probabilities $\Pr(\tilde{r})$, $\Pr(r')$, and CI scores for the continued-counting hypothesis and question-first setting across different values of $K$ and patch types in LLMs.}
\label{tab:l2fllamaQwenQf}
\endgroup
\end{table}

\begin{table}[t] 
\centering

\begingroup
\setlength{\tabcolsep}{2pt} 
\resizebox{0.9\columnwidth}{!}{
\begin{tabular}{@{}llcccccc@{}}
\toprule
 & & \multicolumn{3}{c}{\textbf{Llama3}} & \multicolumn{3}{c}{\textbf{Qwen2.5}} \\
\cmidrule(lr){3-5} \cmidrule(lr){6-8}
\textbf{K} & \textbf{Patch Type} &
$\mathbf{P}(\tilde{r})$ & $\mathbf{P}({r^\prime})$ & \textbf{CI} &
$\mathbf{P}(\tilde{r})$ & $\mathbf{P}({r^\prime})$ & \textbf{CI} \\
\midrule
\multirow{3}{*}{\textbf{1}}
  & Both       & 0.01 & 0.01 & 0.22 & 0.16 & 0.29 & 0.11 \\
  & Separators & 0.03 & 0.47 & 0.00 & 0.05 & 0.40 & 0.00 \\
  & Elements   & 0.01 & 0.01 & 0.22 & 0.16 & 0.29 & 0.11 \\
\cmidrule(l){2-8}
\multirow{3}{*}{\textbf{2}}
  & Both       & 0.00 & 0.00 & 0.21 & 0.10 & 0.10 & 0.17 \\
  & Separators & 0.17 & 0.06 & 0.26 & 0.17 & 0.22 & 0.14 \\
  & Elements   & 0.01 & 0.01 & 0.21 & 0.37 & 0.22 & 0.24 \\
\cmidrule(l){2-8}
\multirow{3}{*}{\textbf{3}}
  & Both       & 0.12 & 0.12 & 0.15 & 0.23 & 0.17 & 0.17 \\
  & Separators & 0.29 & 0.16 & 0.22 & 0.30 & 0.24 & 0.16 \\
  & Elements   & 0.13 & 0.13 & 0.15 & 0.40 & 0.24 & 0.21 \\
\bottomrule
\end{tabular}
}
\caption{Average probabilities $\Pr(\tilde{r})$, $\Pr(r')$, and CI scores for the continued-counting hypothesis and question-last setting across different values of $K$ and patch types in LLMs.}
\label{tab:l2fllamaQwenQl}
\endgroup
\end{table}

\begin{table}[ht]
\centering

\begin{subtable}[t]{0.3\textwidth}
\centering
\resizebox{\columnwidth}{!}{
\begin{tabular}{ccccccccc}
\hline
\textbf{src / trgt} & \textbf{2} & \textbf{3} & \textbf{4} & \textbf{5} & 
\textbf{6} & \textbf{7} & \textbf{8} & \textbf{9} \\
\hline
\textbf{2}  & 2 & 2 & 3 & 4 & 5 & 6 & 7 & 8 \\
\textbf{3}  & 3 & 3 & 3 & 4 & 5 & 6 & 7 & 8 \\
\textbf{4}  & 4 & 4 & 4 & 4 & 5 & 6 & 7 & 8 \\
\textbf{5}  & 5 & 5 & 5 & 5 & 5 & 6 & 7 & 8 \\
\textbf{6}  & 6 & 6 & 6 & 6 & 6 & 6 & 7 & 8 \\
\textbf{7}  & 7 & 7 & 7 & 7 & 7 & 7 & 7 & 8 \\
\textbf{8}  & 8 & 8 & 8 & 8 & 8 & 8 & 8 & 8 \\
\textbf{9}  & 9 & 9 & 9 & 9 & 9 & 9 & 9 & 9 \\
\hline
\end{tabular}
}
\subcaption{$k=1$}
\end{subtable}
\hfill
\begin{subtable}[t]{0.3\textwidth}
\centering
\resizebox{\columnwidth}{!}{
\begin{tabular}{ccccccccc}
\hline
\textbf{src / trgt} & \textbf{2} & \textbf{3} & \textbf{4} & \textbf{5} & 
\textbf{6} & \textbf{7} & \textbf{8} & \textbf{9} \\
\hline
\textbf{2}  & 2 & 2 & 2 & 3 & 4 & 5 & 6 & 7 \\
\textbf{3}  & 3 & 3 & 3 & 3 & 4 & 5 & 6 & 7 \\
\textbf{4}  & 4 & 4 & 4 & 4 & 4 & 5 & 6 & 7 \\
\textbf{5}  & 5 & 5 & 5 & 5 & 5 & 5 & 6 & 7 \\
\textbf{6}  & 6 & 6 & 6 & 6 & 6 & 6 & 6 & 7 \\
\textbf{7}  & 7 & 7 & 7 & 7 & 7 & 7 & 7 & 7 \\
\textbf{8}  & 8 & 8 & 8 & 8 & 8 & 8 & 8 & 8 \\
\textbf{9}  & 9 & 9 & 9 & 9 & 9 & 9 & 9 & 9 \\
\hline
\end{tabular}
}
\subcaption{$k=2$}
\end{subtable}
\hfill
\begin{subtable}[t]{0.3\textwidth}
\centering
\resizebox{\columnwidth}{!}{
\begin{tabular}{ccccccccc}
\hline
\textbf{src / trgt} & \textbf{2} & \textbf{3} & \textbf{4} & \textbf{5} & 
\textbf{6} & \textbf{7} & \textbf{8} & \textbf{9} \\
\hline
\textbf{2}  & 2 & 2 & 2 & 2 & 3 & 4 & 5 & 6 \\
\textbf{3}  & 3 & 3 & 3 & 3 & 3 & 4 & 5 & 6 \\
\textbf{4}  & 4 & 4 & 4 & 4 & 4 & 4 & 5 & 6 \\
\textbf{5}  & 5 & 5 & 5 & 5 & 5 & 5 & 5 & 6 \\
\textbf{6}  & 6 & 6 & 6 & 6 & 6 & 6 & 6 & 6 \\
\textbf{7}  & 7 & 7 & 7 & 7 & 7 & 7 & 7 & 7 \\
\textbf{8}  & 8 & 8 & 8 & 8 & 8 & 8 & 8 & 8 \\
\textbf{9}  & 9 & 9 & 9 & 9 & 9 & 9 & 9 & 9 \\
\hline
\end{tabular}
}
\subcaption{$k=3$}
\end{subtable}

\caption{Expected answer, $\tilde{r}$, under the maximum latent count hypothesis.
The value is computed as
$\tilde{r} = \max\!\big(N_{\text{source}},\, N_{\text{target}} - k\big)$,
where $N_{\text{source}}$ and $N_{\text{target}}$ denote the number of objects in the source and target inputs, respectively, and $k$ is the number of patched items.}
\label{tab:expected:last2last}
\end{table}

The \textit{maximum latent count hypothesis} is tested across both visual and textual tasks. Table \ref{tab:expected:last2last} presents the expected predictions based on this hypothesis for varying values of $k$. Table \ref{tab:l2lInternVLQwen} reports the performance of visual models (InternVL3.5 and Qwen2.5-VL) across different values of $k$ and in both "Question First" and "Question Last" configurations. For textual tasks, the results are shown in Tables \ref{tab:l2lllamaQwenQf} and \ref{tab:l2lllamaQwenQl}, where Llama3 and Qwen2.5 are evaluated with various intervention types (both, separators, and elements). These tables present the average probability of the predicted answer ($\tilde{r}$), the average probability of the incorrect answer ($r^\prime$), and the CI score for each configuration.


\begin{table}[t]
\centering

\begingroup
\setlength{\tabcolsep}{2pt}

\resizebox{0.9\columnwidth}{!}{
\begin{tabular}{@{}llcccccc@{}}
\toprule
 & & \multicolumn{3}{c}{\textbf{InternVL3.5}} & \multicolumn{3}{c}{\textbf{Qwen2.5-VL}} \\
\cmidrule(lr){3-5} \cmidrule(lr){6-8}
\textbf{Question Type} & \textbf{K} &
$\mathbf{P}(\tilde{r})$ & $\mathbf{P}(r^\prime)$ & \textbf{CI} &
$\mathbf{P}(\tilde{r})$ & $\mathbf{P}(r^\prime)$ & \textbf{CI} \\
\midrule
\multirow{3}{*}{\textbf{Question First}}
  & 1 & 0.39 & 0.17 & 0.42 & 0.74 & 0.18 & 0.76 \\
  & 2 & 0.49 & 0.08 & 0.54 & 0.80 & 0.02 & 0.87 \\
  & 3 & 0.58 & 0.04 & 0.61 & 0.86 & 0.01 & 0.90 \\
\cmidrule(l){2-8}
\multirow{3}{*}{\textbf{Question Last}}
  & 1 & 0.57 & 0.25 & 0.53 & 0.64 & 0.25 & 0.66 \\
  & 2 & 0.73 & 0.10 & 0.73 & 0.77 & 0.02 & 0.84 \\
  & 3 & 0.75 & 0.07 & 0.76 & 0.83 & 0.02 & 0.88 \\
\bottomrule
\end{tabular}
}
\caption{ Average probabilities $\Pr(\tilde{r})$, $\Pr(r')$, and CI scores for the maximum latent count hypothesis across different values of $K$ and prompt structures in LVLMs. In the question-first setting, the counting task is specified in the system prompt.}
\label{tab:l2lInternVLQwen}
\endgroup
\end{table}

\begin{table}[t] 
\centering

\begingroup
\setlength{\tabcolsep}{2pt} 
\resizebox{0.9\columnwidth}{!}{
\begin{tabular}{@{}llcccccc@{}}
\toprule
 & & \multicolumn{3}{c}{\textbf{Llama3}} & \multicolumn{3}{c}{\textbf{Qwen2.5}} \\
\cmidrule(lr){3-5} \cmidrule(lr){6-8}
\textbf{K} & \textbf{Patch Type} &
$\mathbf{P}(\tilde{r})$ & $\mathbf{P}({r^\prime})$ & \textbf{CI} &
$\mathbf{P}(\tilde{r})$ & $\mathbf{P}({r^\prime})$ & \textbf{CI} \\
\midrule
\multirow{3}{*}{\textbf{1}}
  & Both       & 0.28 & 0.48 & 0.31 & 0.59 & 0.38 & 0.60 \\
  & Separators & 0.03 & 0.85 & 0.00 & 0.00 & 1.00 & 0.00 \\
  & Elements   & 0.28 & 0.48 & 0.31 & 0.59 & 0.38 & 0.60 \\
\cmidrule(l){2-8}
\multirow{3}{*}{\textbf{2}}
  & Both       & 0.66 & 0.01 & 0.74 & 0.91 & 0.00 & 0.95 \\
  & Separators & 0.49 & 0.16 & 0.58 & 0.34 & 0.44 & 0.45 \\
  & Elements   & 0.26 & 0.42 & 0.34 & 0.52 & 0.43 & 0.54 \\
\cmidrule(l){2-8}
\multirow{3}{*}{\textbf{3}}
  & Both       & 0.80 & 0.01 & 0.82 & 0.95 & 0.00 & 0.97 \\
  & Separators & 0.51 & 0.10 & 0.63 & 0.43 & 0.34 & 0.54 \\
  & Elements   & 0.24 & 0.37 & 0.36 & 0.53 & 0.36 & 0.58 \\
\bottomrule
\end{tabular}
}
\caption{Average probabilities $\Pr(\tilde{r})$, $\Pr(r')$, and CI scores for the maximum latent count hypothesis and question-first setting across different values of $K$ and patch types in LLMs.}
\label{tab:l2lllamaQwenQf}
\endgroup
\end{table}

\begin{table}[t] 
\centering

\begingroup
\setlength{\tabcolsep}{2pt} 
\resizebox{0.9\columnwidth}{!}{
\begin{tabular}{@{}llcccccc@{}}
\toprule
 & & \multicolumn{3}{c}{\textbf{Llama3}} & \multicolumn{3}{c}{\textbf{Qwen2.5}} \\
\cmidrule(lr){3-5} \cmidrule(lr){6-8}
\textbf{K} & \textbf{Patch Type} &
$\mathbf{P}(\tilde{r})$ & $\mathbf{P}(r^\prime)$ & \textbf{CI} &
$\mathbf{P}(\tilde{r})$ & $\mathbf{P}(r^\prime)$ & \textbf{CI} \\
\midrule
\multirow{3}{*}{\textbf{1}}
  & Both       & 0.15 & 0.44 & 0.04 & 0.13 & 0.33 & 0.09 \\
  & Separators & 0.12 & 0.49 & 0.00 & 0.05 & 0.43 & 0.00 \\
  & Elements   & 0.15 & 0.44 & 0.04 & 0.13 & 0.33 & 0.09 \\
\cmidrule(l){2-8}
\multirow{3}{*}{\textbf{2}}
  & Both       & 0.34 & 0.11 & 0.32 & 0.31 & 0.13 & 0.28 \\
  & Separators & 0.42 & 0.17 & 0.33 & 0.32 & 0.17 & 0.26 \\
  & Elements   & 0.11 & 0.38 & 0.07 & 0.13 & 0.30 & 0.11 \\
\cmidrule(l){2-8}
\multirow{3}{*}{\textbf{3}}
  & Both       & 0.31 & 0.08 & 0.32 & 0.33 & 0.11 & 0.30 \\
  & Separators & 0.36 & 0.14 & 0.32 & 0.32 & 0.15 & 0.28 \\
  & Elements   & 0.12 & 0.30 & 0.11 & 0.21 & 0.24 & 0.18 \\
\bottomrule
\end{tabular}
}
\caption{Average probabilities $\Pr(\tilde{r})$, $\Pr(r')$, and CI scores for the maximum latent count hypothesis and question-last setting across different values of $K$ and patch types in LLMs.}
\label{tab:l2lllamaQwenQl}
\endgroup
\end{table}

To assess the generalizability of the linear additivity effect across tasks, we apply position-difference vectors learned (averaged over dataset) from a fruit-counting task to an animal-counting task. Table \ref{tab:linear_additivity_transfer_animals} reports the CI scores and accuracies for this transfer experiment, showing that the additivity pattern holds when applying the learned vectors to the animal counting task, with consistent performance across different position-difference values.

\begin{table}[t]
\centering

\begingroup
\setlength{\tabcolsep}{6pt}

\begin{tabular}{@{}ccc@{}}
\toprule
\textbf{$K$} & \textbf{CI (Animals)} & \textbf{Acc (Animals)} \\
\midrule
1 & $0.56 \pm 0.23$ & $0.54 \pm 0.23$ \\
2 & $0.64 \pm 0.16$ & $0.58 \pm 0.12$ \\
3 & $0.81 \pm 0.18$ & $0.78 \pm 0.20$ \\
4 & $0.87 \pm 0.21$ & $0.81 \pm 0.29$ \\
\midrule
Avg & $0.72 \pm 0.20$ & $0.68 \pm 0.21$ \\
\bottomrule
\end{tabular}

\endgroup

\caption{Linear additivity under task transfer. CI scores and accuracies when position-difference vectors are estimated from a fruit-counting task and then applied to an animal-counting task. We report mean~$\pm$~standard deviation over animal types for different position differences $K$.}
\label{tab:linear_additivity_transfer_animals}

\end{table}

Table \ref{tab:separator_shortcut_details} shows the target probability drop when the activations of the first separator are causally patched across all layers and transferred to subsequent separators, comparing monotypic and polytypic settings for different counts.

\begin{table}[t]
\centering
\begingroup
\setlength{\tabcolsep}{4pt}

\begin{tabular}{@{}llcccccccc@{}}
\toprule
\textbf{Category} & \textbf{3} &
\textbf{4} & \textbf{5} & \textbf{6} & \textbf{7} & \textbf{8} & \textbf{9} \\
\midrule
\multirow{1}{*}{\textbf{Monotypic}}

  & 0.26 & 0.54 & 0.83 & 1.00 & 1.00 & 0.97 & 1.00 \\
\cmidrule(l){2-8}
\multirow{1}{*}{\textbf{Polytypic}}
  & 0.52 & 0.53 & 0.88 & 1.00 & 1.00 & 1.00 & 1.00 \\

\bottomrule
\end{tabular}

\caption{Target Probability Drop for Causal Patching of Separator Activations. The drop in target probability is calculated when the activations of the first separator are patched and transferred to subsequent separators across different counts, in both monotypic and polytypic settings.}
\label{tab:separator_shortcut_details}

\endgroup
\end{table}

\section{Natural Data Analysis}
\label{sec:app_natural}

\begin{figure}[t]
    \centering
    \begin{subfigure}[t]{0.76\linewidth}
        \centering
        \includegraphics[width=\linewidth]{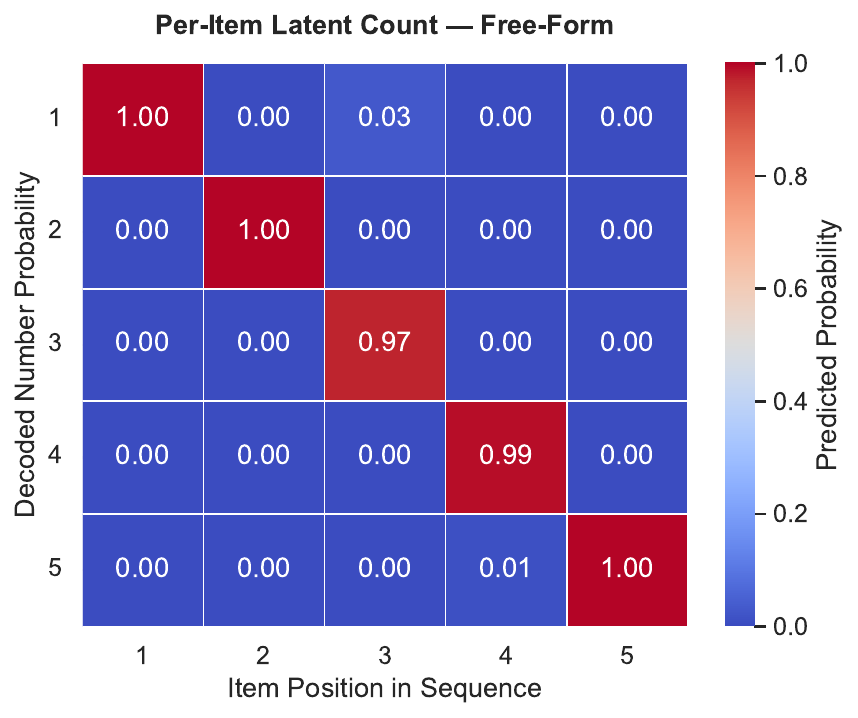}
        \caption{Free-form text}
    \end{subfigure}
    \begin{subfigure}[t]{0.76\linewidth}
        \centering
        \includegraphics[width=\linewidth]{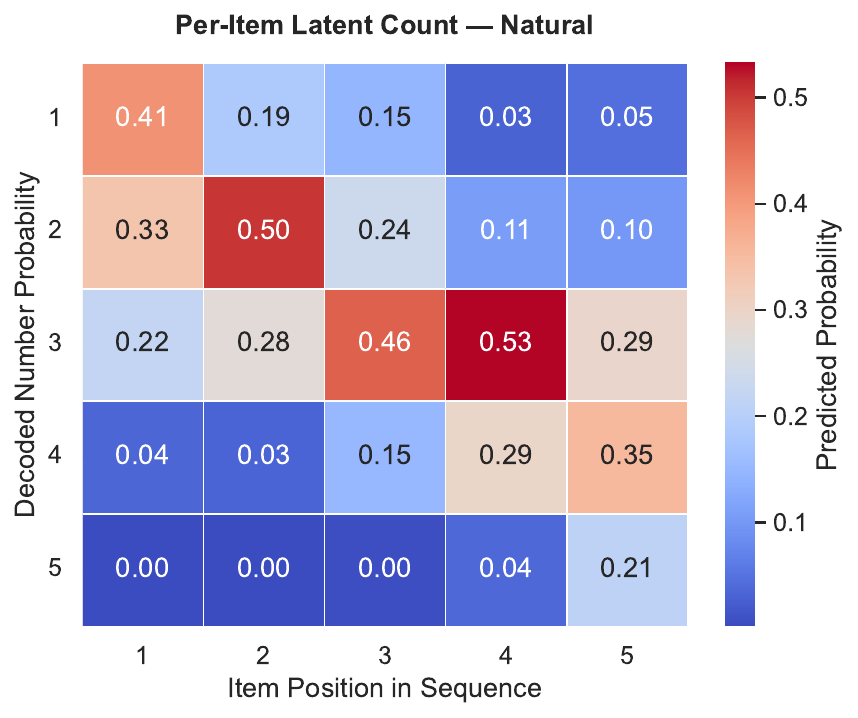}
        \caption{Natural image}
    \end{subfigure}
    \vspace{-2pt}
    \caption{
    Per-item latent count for natural data, decoded by CountScope.
    Each row shows the probability of decoding the count at different sequence positions for separators in (a) unstructured text and (b) real-world images from MS COCO dataset~\cite{mscoco}.
    }
    \vspace{-4pt}
    \label{fig:latentcount_natural}
\end{figure}

In this paper, we focus on synthetic data samples for causal analysis, as they provide a controllable setting and reliable evidence, which is standard in mechanistic interpretability studies. To test generalization beyond synthetic settings, we conduct additional experiments following Section~\ref{sec:internal_counters} for both modalities. For text, we use unstructured contexts with free-form sentence templates containing multiple occurrences of target words. These contexts follow a short naturalistic prose style, such as notes, reports, or transcripts, where repeated target words are embedded in fluent sentences rather than presented as a plain list. A representative template is: ``Short report: Keyword: [WORD]. Repeat: [WORD]. Confirm: [WORD]. Finalize: [WORD]. Close: [WORD].'' where \texttt{[WORD]} is replaced with the name of an object. For vision, we filter real images from the MS COCO dataset~\cite{mscoco} that contain repeated natural object occurrences in diverse scenes.

In both modalities, we apply CountScope to test whether items encode positional count information. We treat item positions as ground truth (GT) and measure the probability assigned to the GT number versus non-GT numbers. For text, the average probabilities for GT and non-GT are 0.81 and 0.02, respectively. For vision, the corresponding probabilities are 0.38 and 0.15, consistent with the standard setting.
Figure~\ref{fig:latentcount_natural} shows the corresponding per-item latent count decoding results for both modalities.
These results indicate that the internal counter mechanism generalizes to free-form text and real-world images.

\begin{figure*}[t]
    \centering
    \begin{subfigure}[b]{0.30\linewidth}
        \centering
        \includegraphics[width=\linewidth]{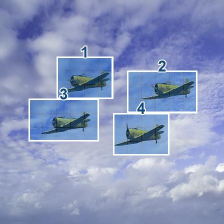}
        \caption{}
    \end{subfigure}
    \begin{subfigure}[b]{0.30\linewidth}
        \centering
        \includegraphics[width=\linewidth]{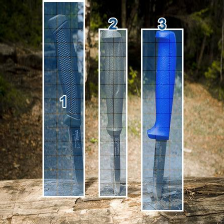}
        \caption{}
    \end{subfigure}
    \begin{subfigure}[b]{0.30\linewidth}
        \centering
        \includegraphics[width=\linewidth]{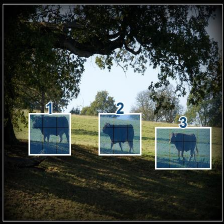}
        \caption{}
    \end{subfigure}
    \begin{subfigure}[b]{0.30\linewidth}
        \centering
        \includegraphics[width=\linewidth]{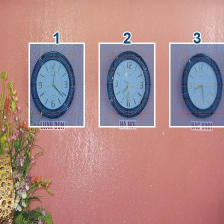}
        \caption{}
    \end{subfigure}
    \begin{subfigure}[b]{0.30\linewidth}
        \centering
        \includegraphics[width=\linewidth]{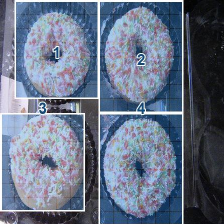}
        \caption{}
    \end{subfigure}
    \begin{subfigure}[b]{0.30\linewidth}
        \centering
        \includegraphics[width=\linewidth]{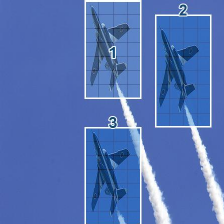}
        \caption{}
    \end{subfigure}

    \caption{
    Qualitative results on natural images.
    Each image contains multiple objects. For each object, its region is patched into CountScope, and the predicted latent count is shown. The overlay heatmap indicates the confidence of the prediction.
    }
    \label{fig:qualitative:synth}
\end{figure*}

We next present qualitative examples of decoded counting for real-world images in Figure~\ref{fig:qualitative:synth}. We observe a consistent pattern in which the model processes and counts objects in a left-to-right and top-to-bottom order. This behavior aligns with the positional encoding of visual tokens and suggests that the internal counter follows a structured spatial traversal of the image.


\section{Layer-Wise Representational Analysis}
\label{sec:app_representational}

\begin{figure*}[t]
\centering
\begin{minipage}{0.48\linewidth}
    \centering
    \includegraphics[width=\linewidth]{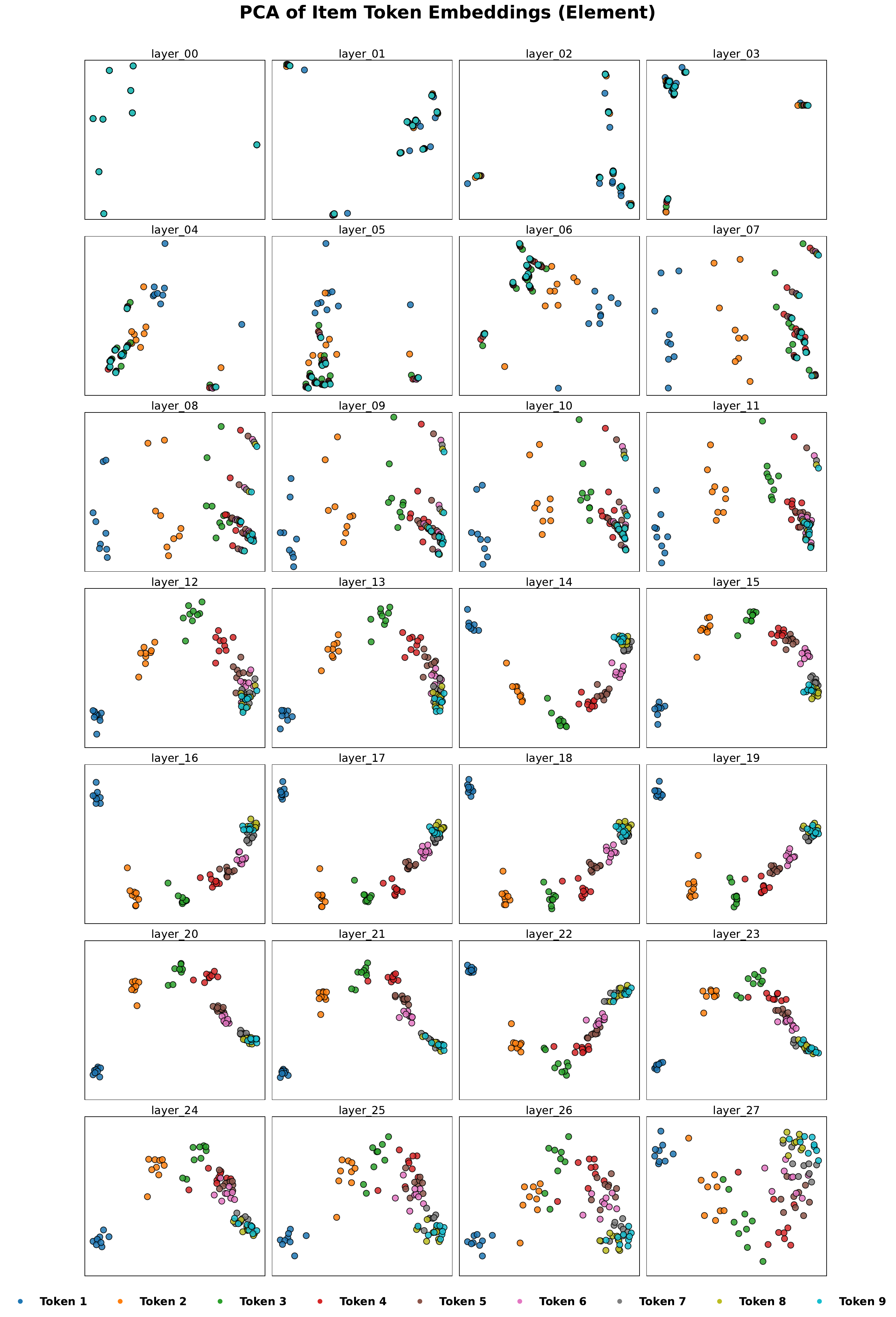}
    \subcaption{PCA of element-token embeddings.}
\end{minipage}
\begin{minipage}{0.48\linewidth}
    \centering
    \includegraphics[width=\linewidth]{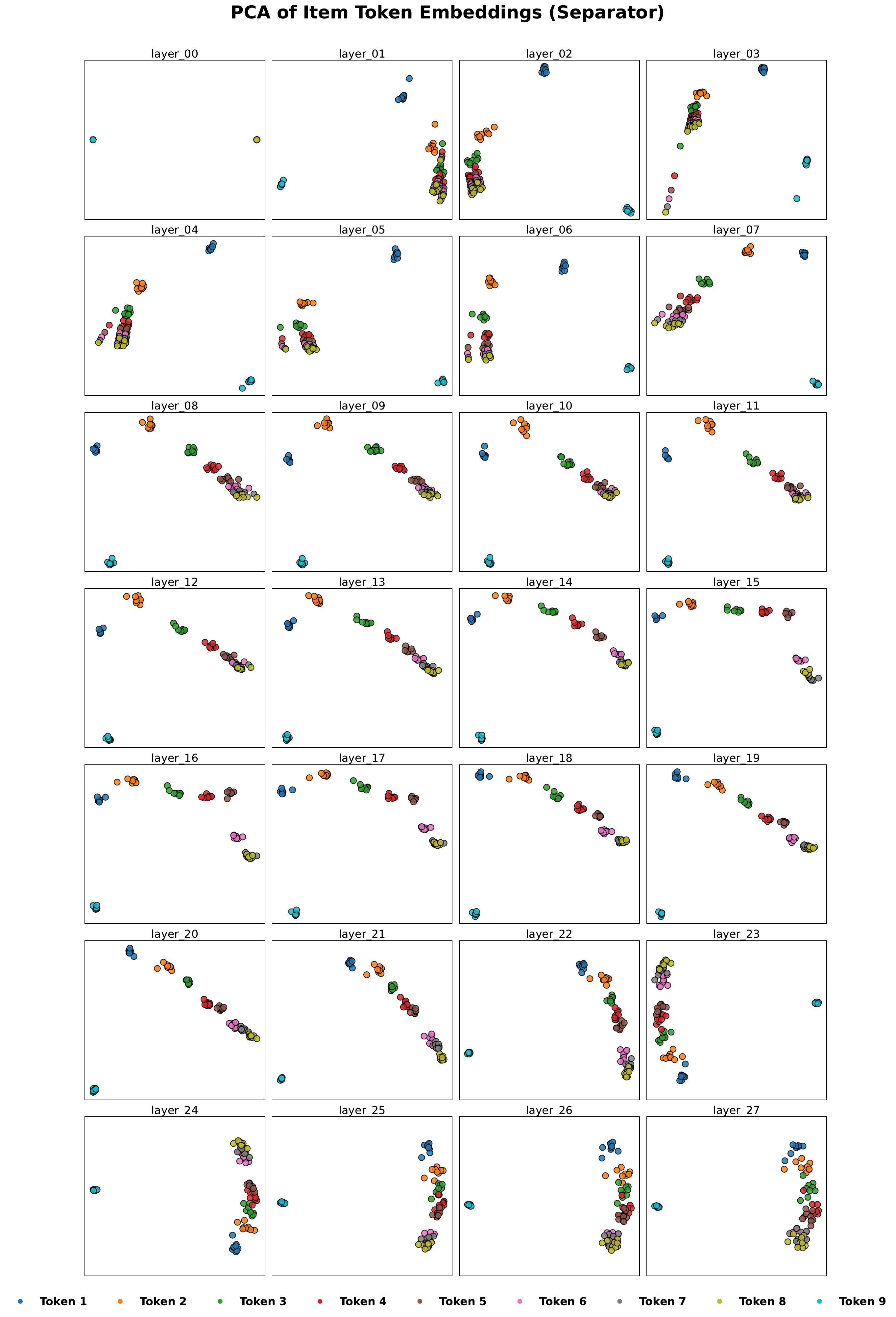}
    \subcaption{PCA of separator-token embeddings.}
\end{minipage}

\caption{
\textbf{Layer-wise PCA of Qwen2.5 input-token representations.}
PCA trajectories across layers for (a) element tokens and (b) separator tokens in the monotypic, question-first setting.
}
\label{fig:layerwise_pca_input}
\end{figure*}

\begin{figure*}[t]
\centering
\begin{minipage}{0.475\linewidth}
    \centering
    \includegraphics[width=\linewidth]{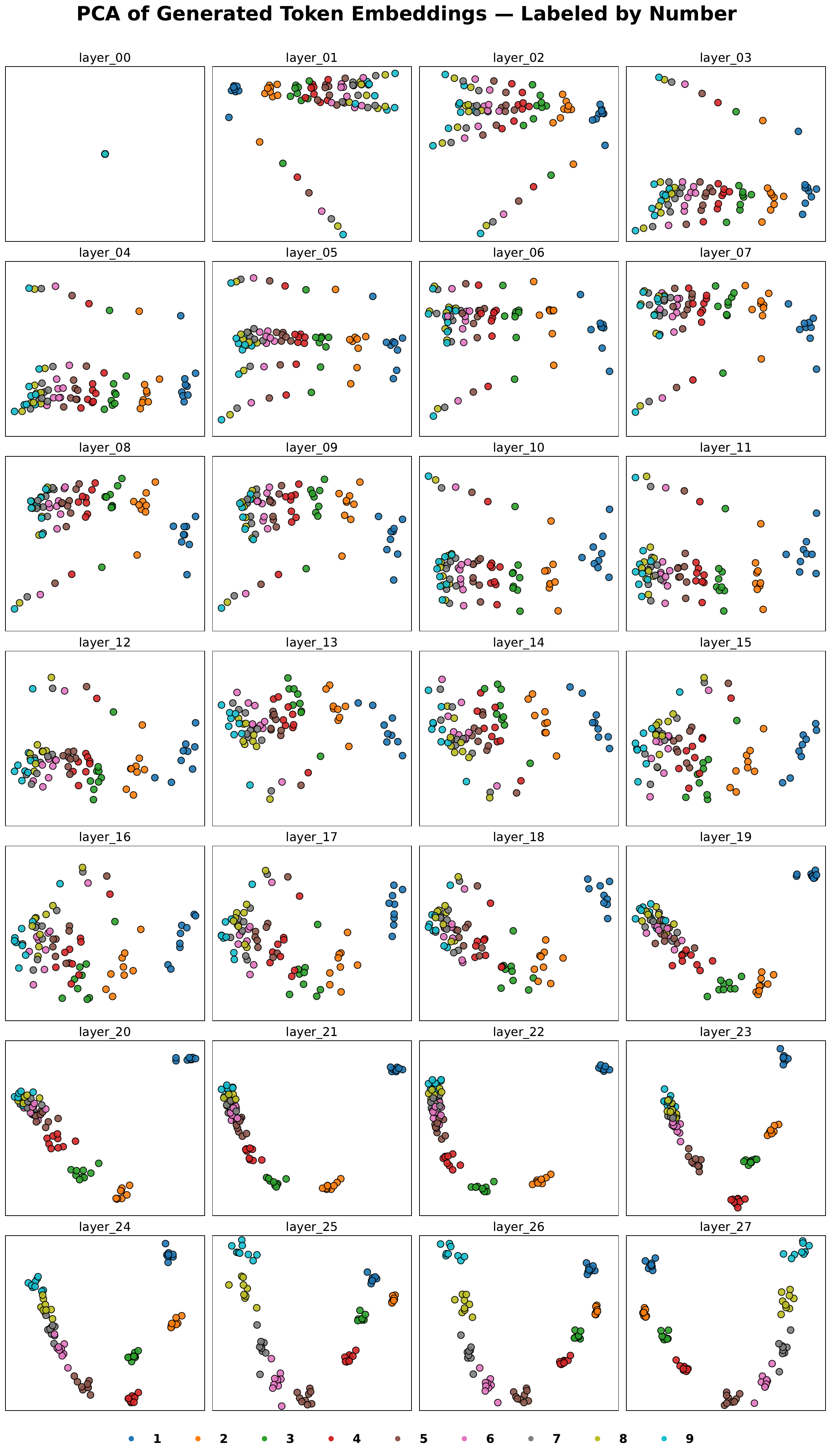}
    \subcaption{PCA of generated responses, colored by predicted number.}
\end{minipage}
\begin{minipage}{0.50\linewidth}
    \centering
    \includegraphics[width=\linewidth]{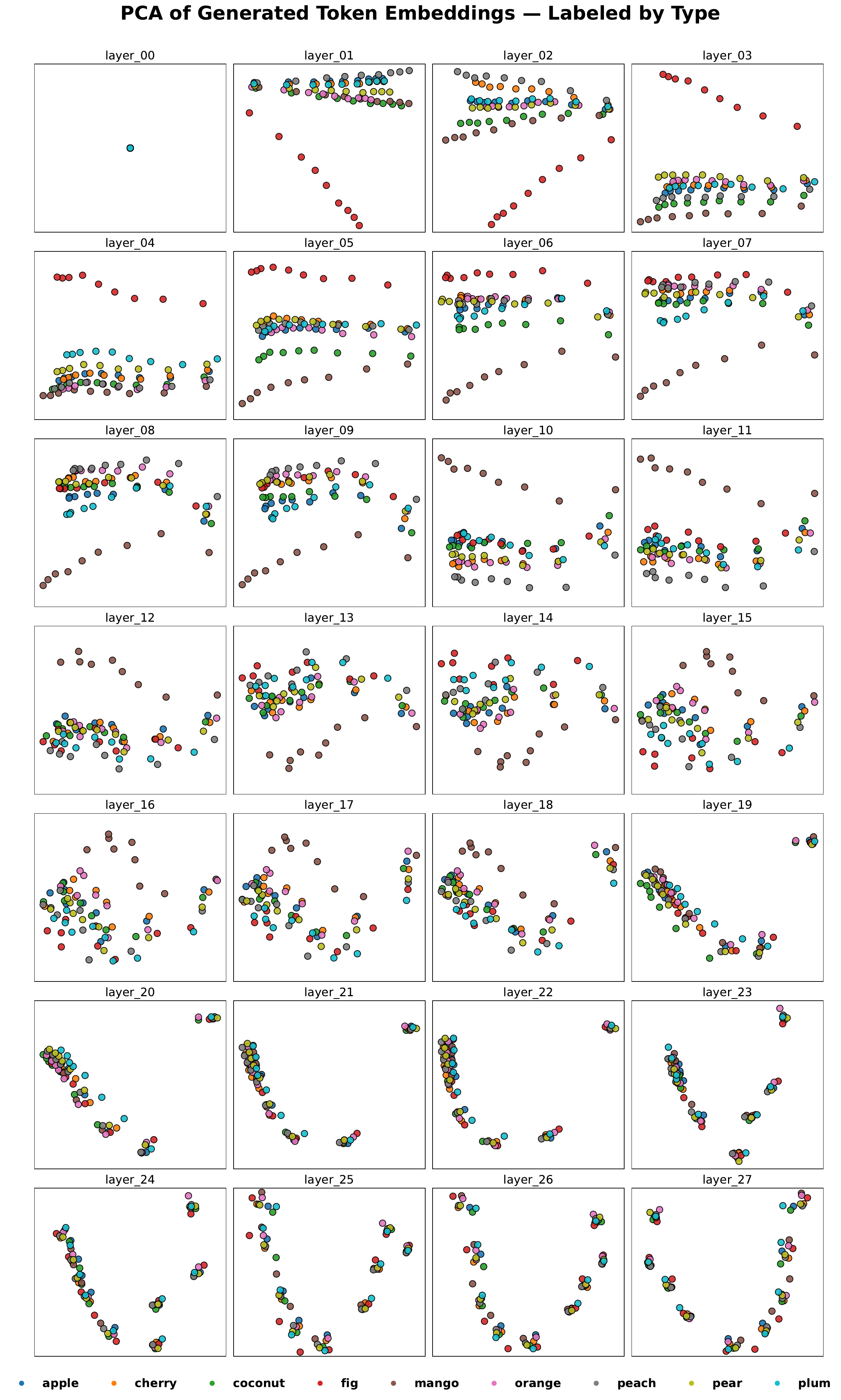}
    \subcaption{PCA of generated responses, colored by item type.}
\end{minipage}

\caption{
\textbf{Layer-wise PCA of Qwen2.5 output representations.}
PCA embeddings across layers for generated numerical responses, colored by (a) predicted count and (b) item type, in the monotypic, question-first setting.
}
\label{fig:layerwise_pca_output}
\end{figure*}

\begin{figure*}[t]
\centering
\begin{minipage}{0.545\linewidth}
    \centering
    \includegraphics[width=\linewidth]{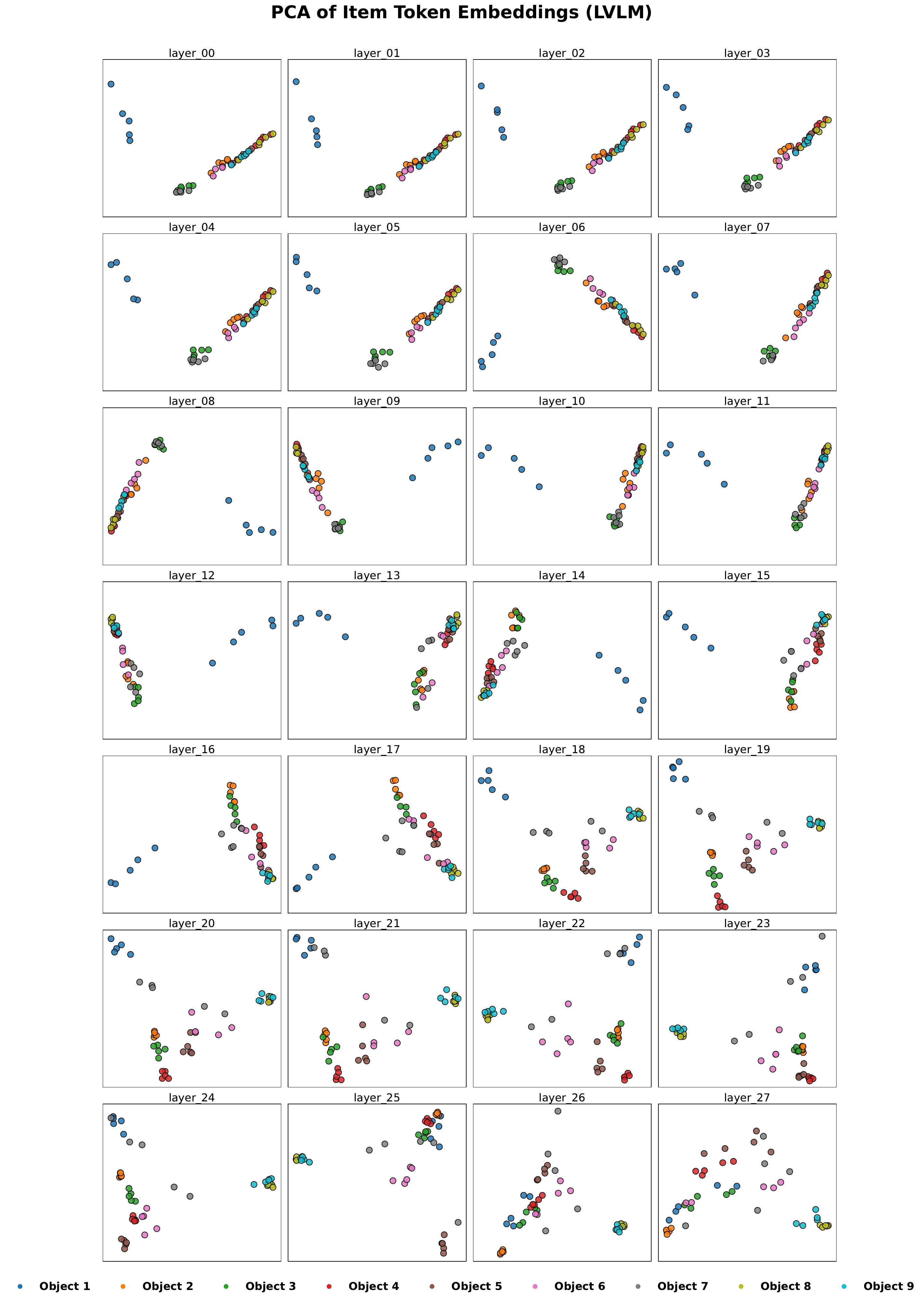}
    \subcaption{PCA of input visual tokens.}
\end{minipage}
\begin{minipage}{0.44\linewidth}
    \centering
    \includegraphics[width=\linewidth]{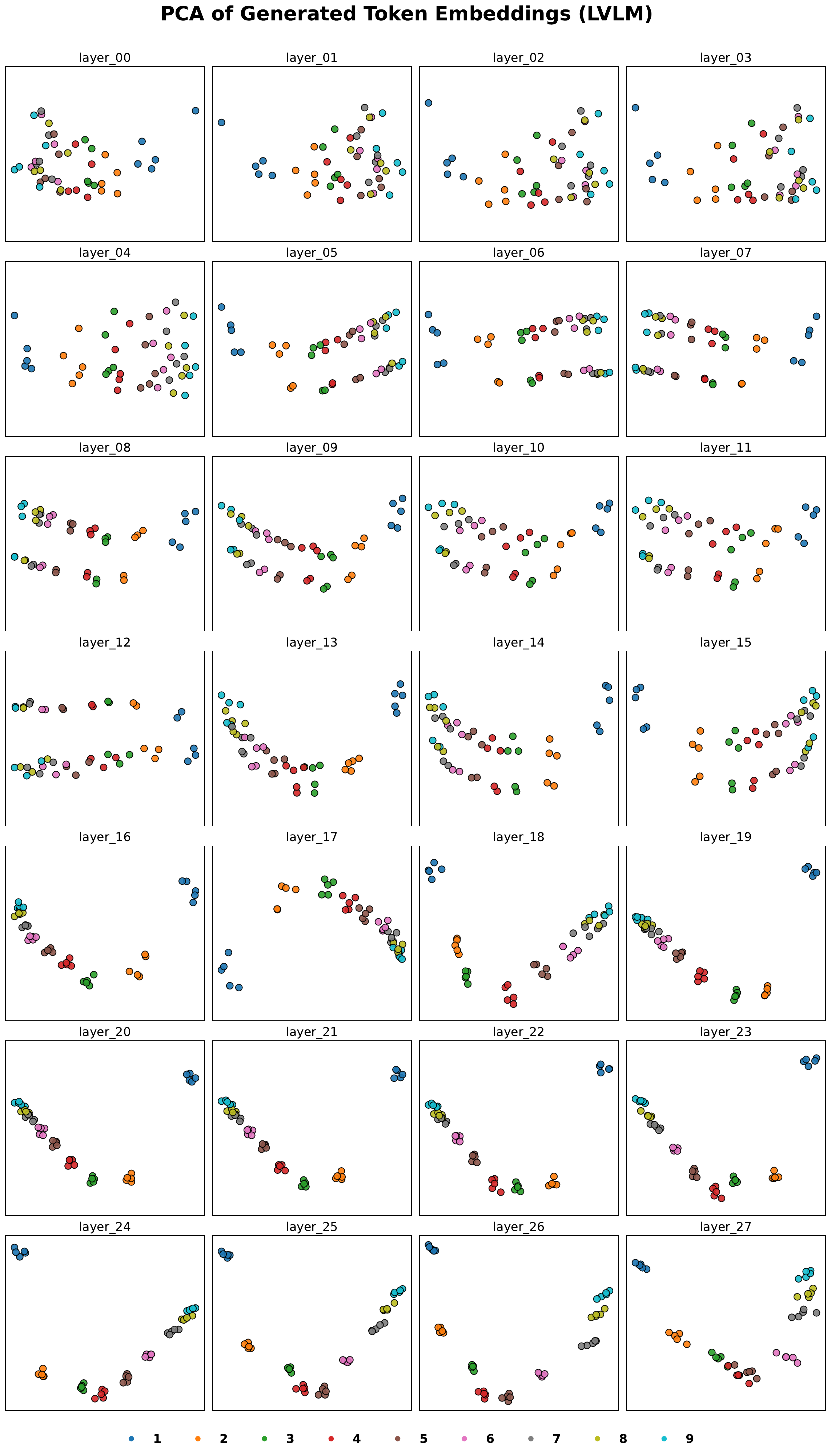}
    \subcaption{PCA of generated response.}
\end{minipage}

\caption{
\textbf{Layer-wise PCA of Qwen2.5VL embeddings.}
PCA trajectories across layers for (a) input item embeddings and (b) generated output embeddings in the monotypic setting.
}
\label{fig:layerwise_pca_lvlm}
\end{figure*}

\begin{figure*}[t]
\centering

\begin{subfigure}[b]{0.48\linewidth}
    \centering
    \includegraphics[width=\linewidth]{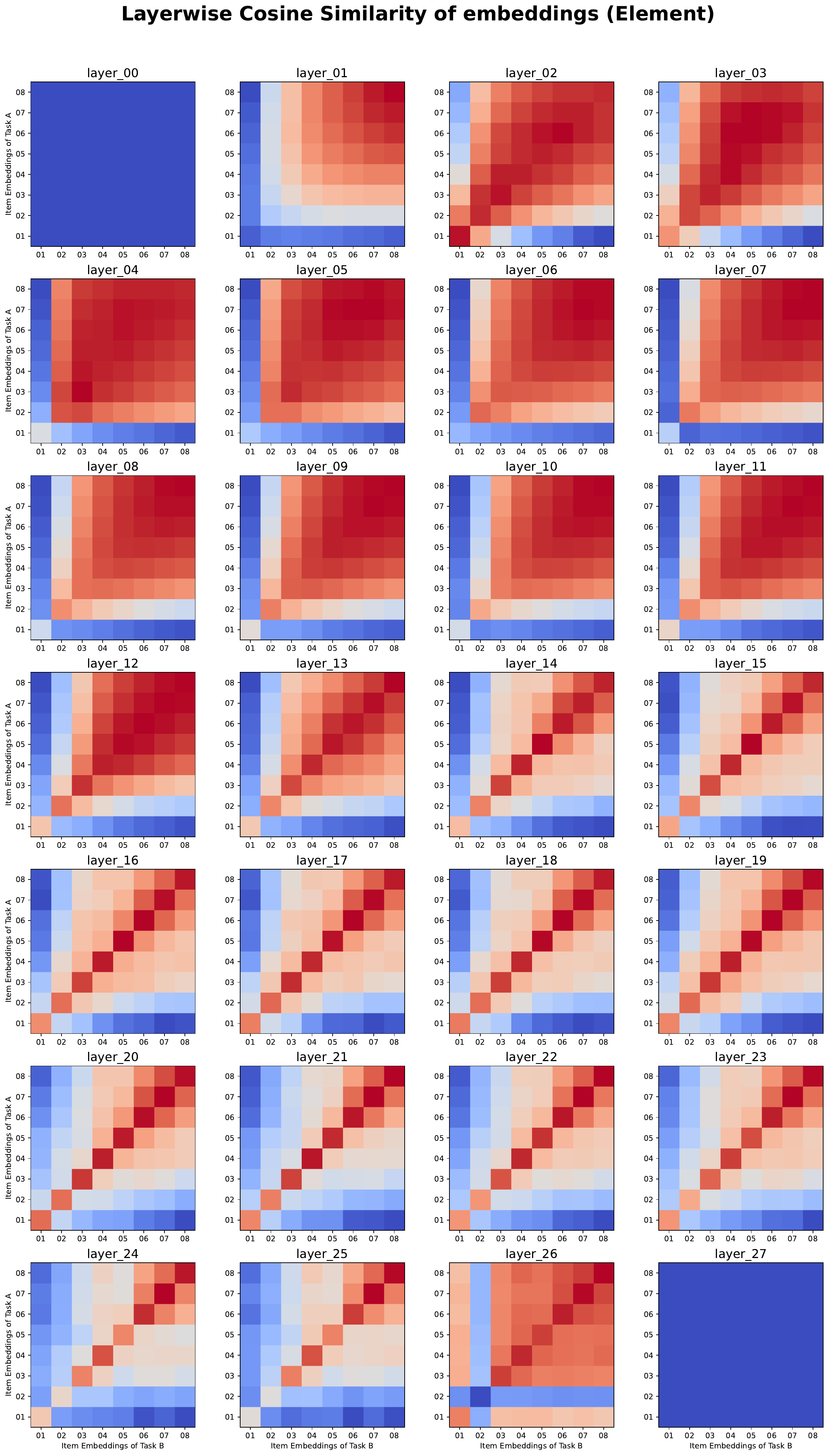}
    \subcaption{Cosine similarity for element tokens.}
\end{subfigure}
\hfill
\begin{subfigure}[b]{0.48\linewidth}
    \centering
    \includegraphics[width=\linewidth]{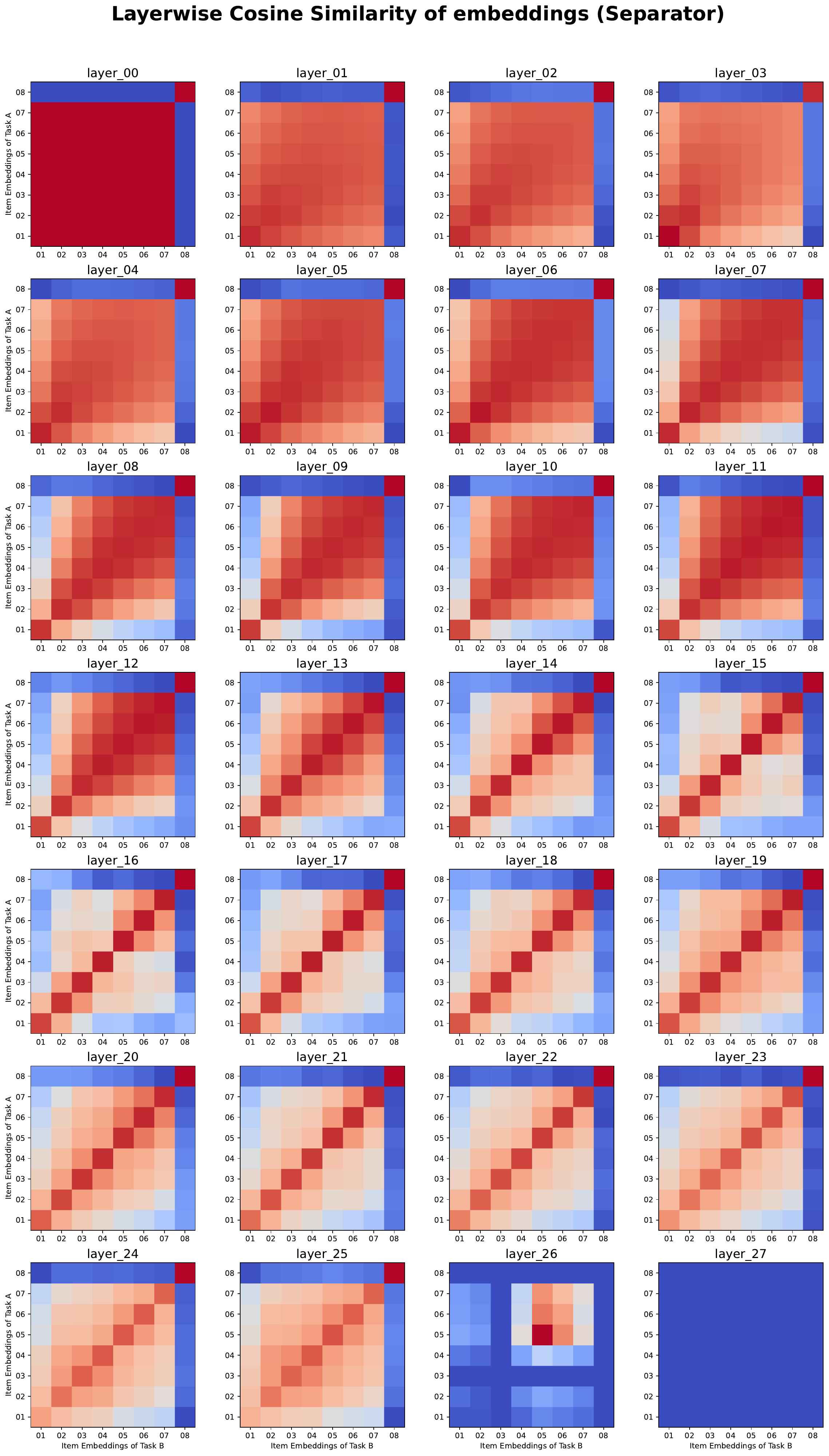}
    \subcaption{Cosine similarity for separator tokens.}
\end{subfigure}

\caption{
\textbf{Layer-wise cosine similarity of Qwen2.5 representations.}
Cosine similarity matrices across layers for (a) element tokens and (b) separator tokens in the monotypic, question-first setting.  
Cosine similarities are computed across different tasks with different item types and then averaged over the dataset.
}
\label{fig:layerwise_cosine_llm}
\end{figure*}

This section provides layer-wise visualizations of representational structure for both LLMs and LVLMs. Figure~\ref{fig:layerwise_pca_input} shows PCA projections of input tokens, and Figure~\ref{fig:layerwise_pca_output} presents PCA of generated responses. 
Figure~\ref{fig:layerwise_pca_lvlm} reports the corresponding trajectories for the LVLM. Figure~\ref{fig:layerwise_cosine_llm} shows cosine similarity patterns across layers for element and separator tokens.

\FloatBarrier

\end{document}